# Understanding the effect of hyperparameter optimization on machine learning models for structure design problems

Xianping Du[1+], Hongyi Xu[2], Feng Zhu[1,3*]

*1 Department of Mechanical Engineering, Embry-Riddle Aeronautical University, Daytona Beach, FL 32114, USA*

*2 Department of Mechanical Engineering, University of Connecticut, Storrs, CT 06269, USA*

*3 Hopkins Extreme Materials Institute and Department of Mechanical Engineering, Johns Hopkins University, Baltimore, MD, 21218, USA*

**Abstract:**

To relieve the computational cost of design evaluations using expensive finite element (FE) simulations, surrogate models have been widely applied in computer-aided engineering design. Machine learning algorithms (MLAs) have been implemented as surrogate models due to their capability of learning the complex interrelations between the design variables and the response from big datasets. Typically, an MLA regression model contains model parameters and hyperparameters. The model parameters are obtained by fitting the training data. Hyperparameters, which govern the model structures and the training processes, are assigned by users before training. There is a lack of systematic studies on the effect of hyperparameters on the accuracy and robustness of the surrogate model. In this work, we proposed to establish a hyperparameter optimization framework to deepen our understanding of the effect. Based on the sequential model-based optimization method, the Pareto front is generated by running the optimal acquisition and updating the surrogate model iteratively. The optimum acquisition works by repeating a design space shrinking process. Using the acquired optimum, the surrogate model is updated, which describes the relationship between the hyperparameter combinations (inputs) generated by Latin hypercube sampling from the design space and structural response (outputs) to evaluate the modeling accuracy. The updated model will then be used for the next iteration of optimal acquisition until the termination criterion is met. Four frequently used MLAs, namely Gaussian Process Regression (GPR), Support Vector Machine (SVM), Random Forest Regression (RFR), and Artificial Neural Network (ANN), are tested on four benchmark examples of structure design optimization. For

---

[+] Current affiliation: Department of Mechanical and Aerospace Engineering, Rutgers University, Piscataway, NJ 08854, USA

[*] Corresponding author: Hopkins Extreme Materials Institute, Johns Hopkins University, 3400 N Charles St, Baltimore, MD 21218, USA
E-mail: fengzhume@gmail.com (Dr. Feng Zhu); Xianping Du (DUX1@my.erau.edu); Hongyi Xu (hongyi.3.xu@uconn.edu)

each MLA model, the model accuracy and robustness before and after the hyperparameters optimization (HOpt) are compared. The results show that HOpt can generally improve the performance of the MLA models in general with dependency on model complexity. HOpt leads to unstable improvements in the MLAs accuracy and robustness for complex problems, which are featured by high-dimensional mixed-variable design space. We also investigated the additional computational costs incurred by HOpt. The training cost is closely related to the MLA architecture. After HOpt, the training cost of ANN and RFR is increased more than that of the GPR and SVM. In summary, this study benefits the selection of HOpt method for different types of design problems based on their complexity (i.e. design domain continuity and the number of design variables, etc.).

## 1. Introduction

In the development of complex structures, such as a vehicle or aircraft, a large number of full-scale numerical simulations are often needed. As the supplement and partial substitute of these expensive simulations, surrogate models have been widely used in engineering design and optimization to reduce the computational cost [1-3]. A surrogate model is established from the design or simulation datasets through regression, to approximate the real model. It provides an efficient way of predicting the responses of new design alternatives without running additional simulations [4]. Surrogate-based methods have been applied successfully in engineering practices, for example, vehicle crashworthiness design [5], crane bridge optimization [6], transportation facility design [7] and so on. To be adapted to various engineering problems, which are characterized by a different number of design variables, degree of nonlinearities, and loading rates and so on, several surrogate models have been proposed, such as polynomial response surface model, radial basis function and Kriging model [5]. However, it is impossible to use one surrogate modeling method to provide accurate predictions for all types of problems [8]. Therefore, much effort was made on the selection of a suitable algorithm for a specific engineering problem [8-10], and the ensemble and aggregation of multiple surrogate models [11, 12] for a higher modeling accuracy.

To comply with the requirement of surrogate modeling with higher accuracy, MLAs have been applied to implement surrogate models for a wide range of problems due to their powerful learning ability and high flexibility [8, 10, 13-18]. The traditional surrogate methods contain only model parameters that can be fitted by data. For example, the constants of the linear regression can be determined by the least square error method; polynomial response surface parameters can be determined by gradient-based methods. Unlike the traditional surrogate modeling approaches, a typical MLA contains not only model parameters but also hyperparameters [19-21], which have to be assigned before model training to control the training process and model structures [22]. They have a great impact on model flexibility, accuracy, and robustness [20, 21, 23, 24].

Up to date, no systematic studies have been conducted regarding the effect of hyperparameters on the MLA-based surrogate modeling in engineering design. Most of the existing studies [25], which applying MLAs as surrogate models, either directly use the default values provided by the MLA package [26], or determine their values based on experience [27] or trial-and-error [28]. Although these surrogate models were trained with good accuracy under pre-determined loss function(s), the potential of the MLAs has not been fully exploited and the way that hyperparameters affect the surrogate modeling performance has not been clarified. To analyze the influence of hyperparameters on the prediction performance, several studies have been performed by parametric analysis in various areas, including the biomechanical analysis [29], material design [30] and mineral exploration [31]. In these studies, however, only a small portion of hyperparameters were tuned from a specific dataset, and the final values of hyperparameters were determined subjectively. Also, the effect of MLA hyperparameters has not been investigated in the area of structural design. Lacking this knowledge, it would be very difficult to fully understand and improve the surrogate modeling of MLAs in the structural design practice. Furthermore, the potential of MLAs cannot be fully exploited solely by traditional parametric studies [32]. The HOpt in structural design is therefore needed. HOpt, which adapts the optimization methods in hyperparameters tuning by taking the hyperparameters as design variables, has been considered as an effective technique to search optimal values of selected hyperparameters. With the help of HOpt, the great improvement of MLAs prediction accuracy has been confirmed based on many public datasets [21, 23, 24, 33]. However, no studies have been reported on HOpt in structure engineering.

In this work, by developing a multi-objective HOpt framework, the effort is made to analyze the surrogate modeling performances of four frequently used MLAs, namely, Gaussian Process Regression (GPR), Support Vector Machine (SVM), Random Forest Regression (RFR), and Artificial Neural Network (ANN). These MLAs are selected as they are widely applied in structural engineering problems, as summarized in Table 1 [11, 27, 28, 31, 34-56]. After a discussion of their accuracy, training cost, and model robustness before and after the HOpt, the hyperparameters tuning of two superior MLAs are also discussed. The method developed in this study can be used to determine the hyperparameters for general structural design problems. The results can help better understanding the effect of HOpt on machine learning models and then be readily extended to solve more complex structural design problems.

The remaining parts of this work are organized as follows. The HOpt framework is described in detail in Section 2 and then it is used to formulate the HOpt for the four MLAs in Section 3. Four engineering benchmark structures are selected and introduced in Section 4, where the simulation datasets are generated. The optimization processes are completed, and the results are presented and analyzed in Section 5 to investigate the hyperparameters effect on modeling accuracy and training cost. A further discussion in Section 6 reveals the effect of HOpt on the surrogate model accuracy and robustness. Besides, parametric studies on the hyperparameters of two superior MLAs, GPR and ANN, are conducted and the results are summarized and documented in the supplement file.

Table 1 Literature on the machine learning algorithms applied as surrogate models in structural engineering

| Literature | Algorithm | Application | Hyperparameters assignment |
|---|---|---|---|
| Mukherjee, A. et al (1995) [34] | ANN | RC beam design | Experience and trail-and-error |
| Kapania, R. K. et al (1998) [35] | ANN | An aerospace continuum beam design | Trail-and-error |
| Nagendra, S. et al (2004) [36] | ANN | A turbine disk performance prediction | Trail-and-error |
| Lee, J. et al (2007) [37] | ANN | A suspension design | Experience and optimization |
| Tang, Y. C. et al (2009) [38] | SVM | The robust design of sheet metal forming process | Experience |
| Guo, Z. et al(2009) [39] | SVM | Reliability analysis for huge space station | Not mentioned |
| Pan, F. et al (2010) [27] | SVM | B-pillar weight minimization using tailor-welded blank (TWB) structure u | Experience |
| Wang, H. et al (2010) [40] | LS-SVM[*a] | The response prediction of a cylinder and whole vehicle crash | Not mentioned |
| Huang, Z. et al (2011) [41] | GPR[*b] | Optimal design of aero-engine turbine disc | Experience |
| Zhu, P. et al (2012) [28] | SVM | Design of vehicle structures for lightweight and crashworthiness | Trail-and-error |
| Zhang, Y. et al (2012) [42] | GPR | Crashworthiness optimization of a foam-filled bitubal square column | Experience |
| Haleem, K. et al (2013) [43] | RFR | To predict the severity of traffic accident | Experience |
| Song, X. et al (2013) [44] | RSM[*c], GPR, SVM, and RBF[*d] | A foam-filled tapered thin-walled structure response prediction | Experience |
| Yin, H. et al (2014) [45] | RSM, RBF, GPR, and SVM | A foam-filled thin-walled structure | Experience |
| Lukaszewicz, D. et al (2014) [46] and (2015) [47] | RFR | Prediction of structure impact performance under manufacturing variation | Not mentioned |
| Rodriguez-Galiano, V. et al (2014) [48] and (2015) [31] | ANN, DT[*c], RFR, and SVM & RFR | Used to map the statistical distribution of mineral prospectivity based on images | Trail-and-error |
| Fang, J. et al (2014) [49] | GPR | To explore the multi-objective design of foam-filled bitubal structures under uncertainty | Trail-and-error |
| Ferreira, W. G. et al (2015) [11] | RSM, GPR, RBNN[*f], and SVM | Analytical and real-world vehicle crashworthiness analysis | Experience |
| Fang, J. et al (2015) [50] | GPR | For the multi-cell tubes optimization | Not mentioned |
| Tang, Z. et al (2016) [51] | RFR | Response prediction of train sets crash with respect to different parameters | Experience |
| Liu, X. et al (2016) [52] | GPR | Demonstrated by a thin-walled box beam and a long cylinder pressure vessel example | Not mentioned |
| Raihan, M. et al (2018) [54] | RFR | Used to find important variables explaining clustered traffic accident data | Not mentioned |
| Duan, L. et al (2018) [55] | SVM | Multi-objectives optimization of a new vehicle longitudinal beam | Experience |
| Palar, P. S. et al (2018) [56] | GPR | The airfoil models design | Optimization |
| Gong, H. R. et al (2018) [57] | RFR | Predicting the international roughness index of asphalt pavements | Trail-and-error |
| Fournier, E. et al (2018) [58] | RFR | Predicting aeronautics loads of a derivative aircraft | Optimization |

Note: [*a] LS-SVM: Least Square-SVM; [*b] GPR is also known as the Kriging method with almost the same basic function. [*c] RSM: Response Surface Method; [*d] RBF: Radial Basis Function; [*e] DT: Decision tree; [*f] RBNN: Radial Basis Neural Network;

## 2. The framework of multi-objective hyperparameters optimization

Four MLAs, i.e. the GPR, SVM, RFR, and ANN, are selected, where the detailed algorithms and model architectures can refer to [59, 60]. They are implemented and trained using the R language package: *mlr* [61] and *mxnet* [62].

The framework with HOpt was established to achieve two basic goals: (1) demonstrate the effect of hyperparameters optimization on the accuracy and robustness of the surrogate models as applied in structural design; and (2) study the effect of each individual hyperparameter on the overall model response. In addition to these functional elements, the framework also realizes a seamless process of structural geometric design, finite element analysis, surrogate modeling and hyperparameter tuning by integrating multiple engineering software.

There are several tools available for HOpt, for example, Hyperopt, Optuna, Tune, scikit-learn, scikit-optimize, SMAC, Autotune, etc. [63, 64] A quantitative comparison with the computational results from these software is essentially a comparison of different optimization algorithms. This has been done in [65, 66] and is not the purpose of the present study. These software tools are focused on MLA performance but not designed and directly used for surrogate modeling in the structural design.

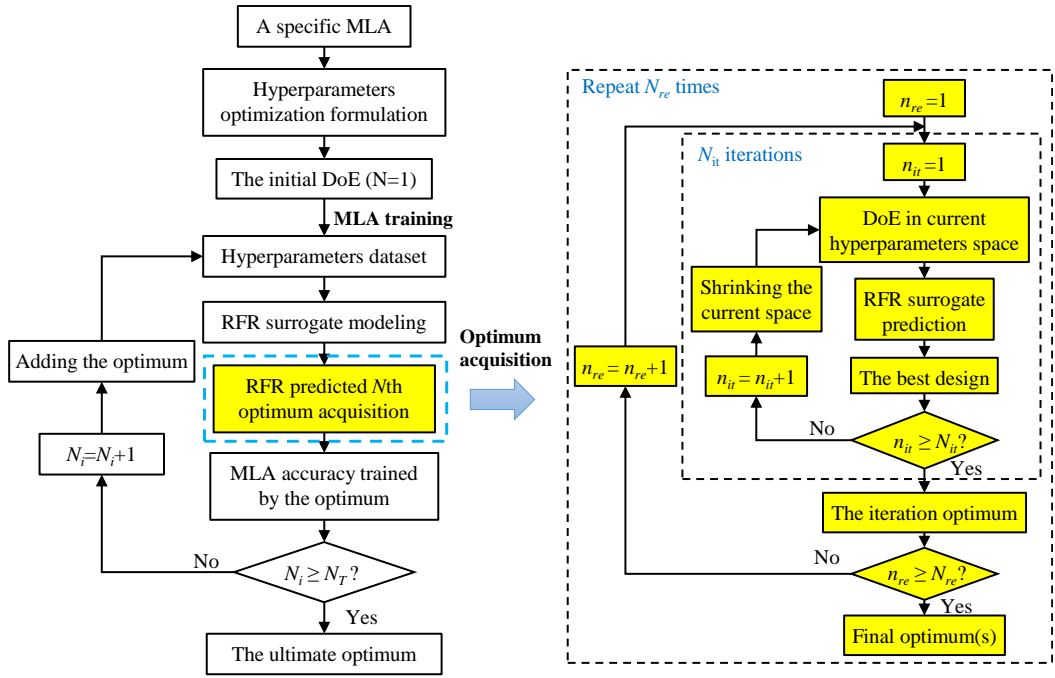

Figure 1 The framework for the multiobjective HOpt based on the sequential model-based method and the detailed workflow of the optimum acquisition

The sequential model-based optimization (SMBO) [19, 67] was used as the basic method as it can reach the optimum efficiently. For a specific MLA, the HOpt problem was formulated first to determine the objective(s) (i.e. searching criteria), tuned hyperparameters, and ranges. The design of experiments (DoE), i.e. multiple combinations of hyperparameters, are generated by Latin hypercube sampling (LHS) function in *lhs* package [68] and evaluated by training the corresponding MLA. The model accuracy quantified by the predefined measures of accuracy is considered as the response of each DoE. Based on this hyperparameters dataset, a surrogate model is trained to represent the relation between the hyperparameters and accuracy measures. In this study, RFR is used as the surrogate model to fit measure values with respect to hyperparameters since it could deal with the categorical features directly, for example, the selection of GPR and SVM kernels and ANN activation functions. The other surrogate methods are less flexible and deal with categorical data by converting to numerical values.

Using the constructed RFR surrogate model, the optimal hyperparameters are predicted and evaluated by training their MLA. The tuning process is terminated when the termination criterion, e.g., the maximum number of iterations ($N_T$), is reached. Otherwise, this optimal design will be added to the hyperparameters dataset and then update the RFR surrogate model for the next HOpt iteration. Finally, a Pareto front can be generated.

In Figure 1, the core part is the optimum acquisition, and its details are introduced as follows. Based on the accuracy measure responses, a number of effective search criteria can be generated as objectives to achieve the trade-off between the exploration and exploitation, for example, the individual or combination of expected improvement (EI), GP upper confidence bound, the maximum possibility of improvement, minimum conditional entropy and the lower confidence bound (LCB) [23, 24, 69]. In this study, the LCB is used, and it is defined as,

$$LCB(x_{hp}, \varphi) = \hat{\mu}(x_{hp}) - \varphi \hat{s}(x_{hp}),$$
(1)

where $x_{hp}$ is the vector of hyperparameters and the $\hat{\mu}(x_{hp})$ and $\hat{s}(x_{hp})$ are the posterior mean and standard deviation of accuracy measures, respectively, from the five cross-validations. $\varphi$ is a constant to count the weight of $\hat{s}(x_{hp})$, where $\varphi = 1$ is used for MLA with all numerical hyperparameters and $\varphi$ equals 2 if at least one hyperparameter is categorical. This counts the possible larger variation caused by categorical hyperparameter(s). $\hat{\mu}(x_{hp})$ will be estimated by the RFR surrogate output directly, but $\hat{s}(x_{hp})$ should be approximated by multi-responses generated by bagging technique in RFR. This criterion demonstrates the lower bound of the accuracy measures reflects the error induced by RFR prediction ($\hat{s}(x_{hp})$).

Using LCBs as objectives, the multi-objective HOpt will be implemented by using a design space shrinking procedure recursively. The design space shrinking refines the design space iteratively for reaching the final optimum or satisfying design. Khan et al (2019) proposed an interactive generative design system for yacht hull by including the design space shrinking, space-filling sampling [70], and human evaluator [71]. The genetic algorithm was also implemented to enhance optimal searching [72]. However, the intensive involvement of manual evaluation by a human may increase the degree of uncertainty. Thus, the surrogate models were developed to replace the evaluator and realize a semi-auto process using ML tools, e.g., the ANN [73]. In the present work, the RFR method is adapted as the surrogate model for the response of design alternatives to enable an automatic process to seek the optimum. Compared with other regression methods, RFR can handle the categorical parameters more efficiently with a relatively lower computational cost. In addition, the space shrinking process was repeated multiple times with different initial designs to increase the probability of reaching the true optimum.

In the framework, within a single iteration, a group of designs (hyperparameters combinations) will be sampled in the design space of this iteration by Latin Hypercube Sampling (LHS), which aims to spread a group of samples almost uniformly over the design space [74]. The response of design alternatives is predicted by the RFR surrogate model. The best alternative will be selected and a smaller subspace will be generated around this point. The construction of the new subspace needs to consider both exploration and exploitation. A small subspace improves convergence speed but may lose the optimum. Similarly, a larger subspace may seize the optimum but causes a low convergence speed. In this study, the ($i+1$)th subspace is defined as $x_{hp\_i} \cap X^*_{hp\_i} \pm r_p \cdot |x_{hp\_i}|$, where $x_{hp\_i}$ and $|x_{hp\_i}|$ represent the hyperparameters space of the $i$th iteration and its sizes and $x^*_{hp\_i}$ is the optimum of $i$th iteration. $\cap$ and $r_p$ are the intersection operation of two subspaces and a proportion to define the expanded space size centered on the best design of the current iteration. In this smaller subspace, a new round of sampling and evaluation process will be implemented until the termination criterion is reached. The total number of the space shrinking iteration is limited to $n_{it}$. After the iterations are terminated, an optimum design can be obtained.

To fully explore the design space and improve robustness, this optimization process will be repeated $n_{re}$ times. Then, a group of designs is generated by taking the Pareto front from each of the $n_{re}$ optimization processes, where each process is implemented with $n_{it}$ iterations for space shrinking as described above. The overall Pareto front will be obtained from this group of designs and used to determine the hyperparameter values of corresponding MLA models.

## 3. Implementation of hyperparameters optimization

Based on available datasets, the hyperparameters can be optimized by the above framework. Before this, the hyperparameters domain, measures, and formulation should be implemented.

### 3.1 Design synthesis and post-processing

The dataset used for MLA training can often be generated by the design of experiment (DoE). Many algorithms are available to create DoE, e.g. Pseudo-Monte Carlo Sampling (PMCS), Latin Hypercube Sampling (LHS), and Orthogonal Array Sampling (OAS) [75]. In this study, the LHS is also used since it could fill the design space uniformly to well explore the design space.

The size of the training dataset must be sufficiently large to ensure convergence and modeling accuracy. Yang et al (2005) [76] and Shi et al (2012) [10] determined the $3 V_N$ is the minimum sample size to train a good surrogate model, where $V_N$ is the number of design variables. Furthermore, Xu et al (2016) [77] developed a polynomial coefficient metric to evaluate the adequacy of sample size considering the different degrees of the nonlinearity of design problems. Based on these theories, 1,000 data points are considered enough in the present case.

It is also noted that design variables may have different orders of magnitudes. This would outweigh features with a larger value over these with smaller value [78]. Therefore, the min-max normalization is used to scale the values of design variables into the same range [0, 1] by

$$NV = \frac{v - v_{\min}}{v_{\max} - v_{\min}}, \qquad (2)$$

where for a specific variable, the $NV$ is the normalized value, $v$ is the variable value; $v_{\max}$ and $v_{\min}$ are the maximum and minimum value of this variable, respectively.

### 3.2 Hyperparameters domains

For each MLA, hyperparameters are identified for optimization considering their significant impact on model accuracy. Their initial values and ranges are determined based on the experience and previous studies as presented in Table 2. For GPR, the kernels and related parameters are tuned as the bold parameters in Equation (3),

$$\boldsymbol{Kernels}: \begin{cases} \text{rbfdot (radial basis):} & k(x, x') = \exp(-\boldsymbol{\sigma} \|x - x'\|^2) \\ \text{polydot (polynomial):} & k(x, x') = (\boldsymbol{scale} <x, x'> + \boldsymbol{offset})^{\boldsymbol{deg}} \\ \text{tanhdot (hyperbolic tangent):} & k(x, x') = \tanh(\boldsymbol{scale} <x, x'> + \boldsymbol{offset}) \\ \text{laplacedot (Laplacian):} & k(x, x') = \exp(-\boldsymbol{\sigma} \|x - x'\|) \end{cases}, \qquad (3)$$

where, $\|x - x'\|$ is the Euclidean distance of vector $x$ and $x'$; $\boldsymbol{scale}$ and $\boldsymbol{offset}$ are used to scale the result of $<x, x'>$ and add an offset, respectively. $\boldsymbol{deg}$ defines the degree of the polynomial kernel. $\sigma$ is the weight of new nodes to the training nodes. The kernel function determines the ability to model the complexity and nonlinearity of the structural problem.

Besides kernel parameters, the control parameters for model complexity and accuracy are tuned for SVM. SVM maps design variables to a new space with higher ($h$) dimensions for a linear regression as expressed by,

$$f(\boldsymbol{x}) = \boldsymbol{\omega}^T \mu(\boldsymbol{x}) + b, \qquad (4)$$

where $\mu(x)$ is the mapping function: $\boldsymbol{x} \rightarrow \mu(\boldsymbol{x}) \in \mathbf{R}^h$. The training objective compromises model complexity and accuracy by Equation (5). The complexity is controlled by the first term, namely, $\frac{1}{2}\boldsymbol{\omega}^T\boldsymbol{\omega}$ [79, 80]. The second term manages the error by the Vapnik's $\varepsilon$ - intensive cost function to penalize the data points outside $\varepsilon$ -bands [81] under the constraints in Equation (6).

$$min \quad \frac{1}{2}\boldsymbol{\omega}^T\boldsymbol{\omega} + C\sum_{i=1}^{N}(\boldsymbol{\xi}+\boldsymbol{\xi}^*) \tag{5}$$

subject to:

$$\begin{cases} y_i - (\boldsymbol{\omega}^T\mu(\boldsymbol{x}_i)+b) \leq \varepsilon + \xi_i \\ (\boldsymbol{\omega}^T\mu(\boldsymbol{x}_i)+b) - y_i \leq \varepsilon + \xi_i^* \\ \xi_i, \qquad \xi_i^* \qquad \geq 0 \end{cases}, \tag{6}$$

where $C$ is the weight of error penalization. $\xi_i$ and $\xi_i^*$ are two slack variables, introduced by Cortes and Vapnik et al. [81], to take points out of the $\varepsilon$ -bounds back to constraints. The Lagrange method is used by introducing constraints into the objective through Lagrange multiplier $\alpha$ and $\alpha^*$ [79, 81] and solving a dual problem by the quadratic programming procedure.

By tunning the $C$ and $\varepsilon$, the SVM training can reach a trade-off between the model complexity and accuracy. A smaller $\varepsilon$ will add more items to the latter term in Equation (5) and larger $C$ will overweight error terms, which may cause the highly complex objective function. These may reach a highly accurate SVM model but take the risk of overfitting since the relative underweighting of the former term in Equation (5) may cause a too complex model and so versa. The HOpt aims to gain the trade-off between model complexity and accuracy.

In RFR, the number of regression trees is a key factor for accuracy, but it is also critical to the computational cost. With more trees included, a higher computational cost will be caused. Meanwhile, the tree-related parameters are also optimized, i.e., the number of randomly selected features for each split (NF), minimum terminal node size (Min TS) and maximum numbers of terminal nodes (Max TN).

In the structural design, it is quite often that more layers of ANN will significantly increase the computational time but not necessarily improve the modeling performance [82]. Based on our preliminary studies shown in Table 1, a wide range of structures with 5~15 design variables can always be well modeled using one-layer ANN. This finding is consistent with Kolmogorov theorem [83-85], that any continuous real-valued functions $f(x_1, x_2, \cdots, x_n)$ defined on [0, 1] can be represented with ANN with one hidden layer. More layers did not increase the accuracy but tended to cause overfitting as shown in the supplement material. A deep network would be more suited to model the systems with deep features, e.g., vibration signal processing [86], natural language processing and image processing [87], but not a typical structural design problem. Thus, the ANN structure with single hidden layer is selected.

Table 2 Hyperparameters of four MLAs with their initial values for the benchmark models and tuning space for optimization

| MLA | Items | Hyperparameters | | | | | | |
|-----|-------|-----|-----|-----|-----|-----|-----|-----|
| | | Kernels | | | σ | Degree | scale | offset |
| GPR | Initial value | rbfdot | | | 0.5 | -- | -- | -- |
| | Range | <rbfdot, polydot, tanhdot, laplacedot> | | | [0 10]*a | 1:10*b | [0 10] | [-10 10] |
| | | C | ε | Kernels | σ | Degree | scale | offset |
| SVM | Initial value | 1 | 0.1 | rbfdot | 0.5 | -- | -- | -- |
| | Range | [0 10] | [0 1] | <rbfdot, polydot, tanhdot, laplacedot>*c | [0 10] | 1:10 | [0 10] | [-10 10] |
| | | Trees | NF | Min TS | | Max TN | | |
| RFR | Initial value | 500 | 3 | 5 | | Null*d | | |
| | Range | 1:1,000 | 1:100 | 1:50 | | 1:1,000 | | |
| | | Hidden neurons | Activation | Optimizer | Batch size | Learning rate | Momentum | |
| ANN | Initial value | 10 | tanhdot | sgd | 120 | 0.1 | 0.0 | |
| | Range | 1:100 | <tanhdot, relu, sigmoid, softrelu> | <sgd, rmsprop, adam, adagrad> | 50:200 | [0.01 1.0] | [0.5 0.99] | |

Note: *a [1 10] means the continuous variables with a range from 1 to 10; *b 1: 10 means the integer variable with values from 1 to 10; *c < rbfdot, polydot, tanhdot, laplacedot> represents a categorical variable with four options available in the curly braces. *d Null means there is no limitation for this hyperparameter and depends on the requirements of other hyperparameters.

The ANN structure and training related parameters are selected for optimization, that is, the number of hidden neurons, mini-batch sizes, activation functions, optimizers, learning rate, and momentum, where the momentum is only applicable to the *sgd* optimizer. ANN training aims to optimize weights for a predefined architecture. To increase this number may improve the accuracy but take the risk of overfitting. The mini-batch method trains ANN with a randomly selected portion of training dataset, which divides the training dataset into several subsets, and each will be used to train the ANN in sequence within a single epoch. This speeds up the convergence of the stochastic convex optimization but increases the training cost of a single epoch. Optimizers determine the strategies to move towards the optimum. The *sgd* (stochastic gradient descent) updates weights in the form of

$$\boldsymbol{w}_{t+1} = \boldsymbol{w}_t - \zeta \cdot \nabla \boldsymbol{w}_t + \xi \cdot \Delta \boldsymbol{w}_{t-1}, \tag{7}$$

where $\boldsymbol{w}_t$ and $\nabla \boldsymbol{w}_t$ are the weight matric and gradient of the current epoch, respectively, and $\Delta \boldsymbol{w}_{t-1}$ is the step size of last epoch. $\zeta$ and $\xi$ are the learning rate and momentum, respectively. A large learning rate accelerates the convergence speed but may lose the optimum while the too small value causes the high convergence cost. For other optimizers, adaptive learning rates are also used to speed up the learning process [87]. In this study, ANN training epochs were limited to 2,000, which is proven to be enough for training convergence by an additional test.

### 3.3 Measures of accuracy

During the surrogate modeling, a number of statistical indices, such as the root mean square error (RMSE), maximum absolute error (MXAE), mean absolute error, correlation, and decision coefficient, are frequently used as measures of accuracy [17, 88, 89]. Although there is no physical meaning, they are calculated based on the quantities obtained in the simulations/experiments with physical meaning, such as force, stress, deformation, and energy absorption, etc. These quantities are often design objectives and may vary in each particular design problem. To keep consistent in all of the structural design scenarios in the present work and with the studies by other researchers, In this study, the RMSE (Equation (8)) and MXAE (Equation (9)) are used as two HOpt objective measures [88] as the global and local accuracy measures, respectively, to estimate the modeling accuracy.

$$RMSE = \sqrt{\frac{\sum_{i=1}^{N} (\tilde{f}(x^i) - f(x^i))}{N}}, \tag{8}$$

$$MXAE = \text{Max}\left( \left| \tilde{f}(x^i) - f(x^i) \right| \right), \tag{9}$$

where $\tilde{f}(x^i) - f(x^i)$ is the error of the predicted response relative to its real value of the *i*th design ($x^i$); *N* is the dataset size. Also, the training computational time (*T*) of MLAs is evaluated. This helps understand the effect of HOpt on the cost and make a decision on the selection of a suitable MLA in terms of computational cost.

### 3.4 Optimization formulation

Based on the hyperparameters and measures, the optimization problem is formulated in Equation (10). RMSE and MXAE are taken as the objectives of the HOpt. For each HOpt task of MLA, 30 initial designs are generated to construct the initial surrogate model. Another 70 evaluations (i.e., iterations, $N_T$ =100) are used to optimize the hyperparameters. $N_T$=100 is verified to be sufficient to ensure the convergence of Pareto front by a separate parametric analysis with a reasonable computational cost as shown in Appendix A. The 5-folds cross-validation is used to train the model, which divides the training dataset into five subsets to train the model in iterations. Each subset is used for validation and the rest are used to train a new model with five validation errors generated. The average of the validation errors from the 5-folds is used as the final validation error. The cross-validation approach is widely used and could reduce the bias more effectively compared to the traditional single train/validation split method. Since bias is considered as the cause of overfitting, the cross-validation approach tends to reduce overfitting [87, 90, 91].

Find: $\boldsymbol{x}_{hp}$

Minimize: *RMSE* and *MXAE*

Subject to: $\begin{cases} \boldsymbol{x}_{hp\_I} \in \min(\boldsymbol{x}_{hp\_I}) : \max(\boldsymbol{x}_{hp\_I}) \\ \min(\boldsymbol{x}_{hp\_C}) \leq \boldsymbol{x}_{hp\_C} \leq \max(\boldsymbol{x}_{hp\_C}) \\ \boldsymbol{x}_{hp\_Ca} \in <x_{hp\_Ca}^1, x_{hp\_Ca}^2, \cdots x_{hp\_Ca}^{NCa}> \end{cases}$ (10)

where $\boldsymbol{x}_{hp\_I}$ $\boldsymbol{x}_{hp\_C}$ $\boldsymbol{x}_{hp\_Ca}$ are the integral, continuous and categorical variables, respectively. Their domains, possible values or ranges, have been defined in Table 2. The datasets generated by DoE are used for MLAs training since the final aim is to construct surrogate models that predict the structural response(s) accurately.

## 4. Benchmark problems for the study of MLAs

In this study, four representative engineering structures are taken as examples to demonstrate the effect of hyperparameters by the HOpt. In these four problems, the variation of geometry (bar, sheet, and block), loading (static vs. dynamic) and boundary conditions (fully constrained and contact), as well as deformation modes (small vs. large deformation) are all considered. Hence, the methods and results associated with these case studies can be easily extended to a wide range of structures.

### 4.1 Structures subject to static loading

Under the static loading condition, two typical structures, i.e. a ten-bar planer truss (TbPT) [32] and a torque arm (TqA) [92-95], are introduced in Figure 2, since they are frequently taken as examples to verify structural design algorithms. In the TbPT model, the circular cross-section areas are taken as design variables with the range from 0.6 to 225.8 cm² and the other sizes are listed in Figure 2(a). Two loads with the same magnitude (444.8 kN) are applied on joints **T** and **S**. The *vertical displacement* (*d*) of joint **S** is taken as the response of this system, which is constrained within 60 cm in case the failure risk caused by the too-small cross-section areas.

To design the TqA structure, geometric constraints must be considered. The constant thickness of this structure is 3 mm and the distance between two circle centers is 420 mm. The design variables and their ranges are listed in Table 3. The left circle is fully constrained and loads are added to the right circle as shown in Figure 2(b). To avoid structure failure, the stress is limited under 800 MPa and the *total mass* is used as the objective.

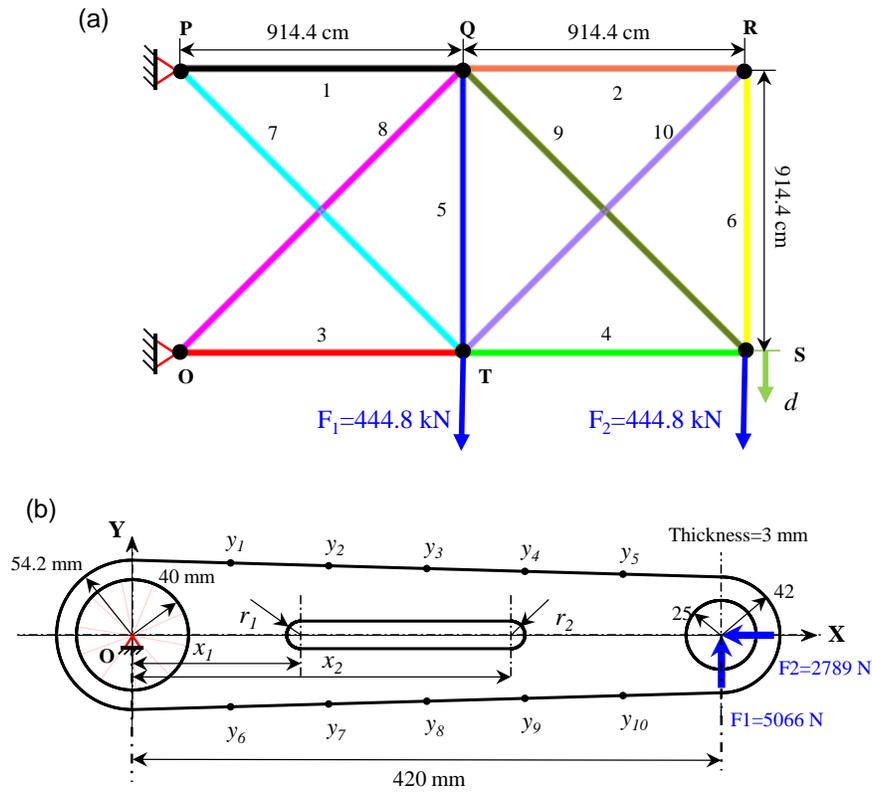

Figure 2 Two structures under static loading: (a) TbPT and (b) TqA

Table 3 Design variables and their ranges for the TqA structure

| Design variable | Initial value /mm | Range/mm | Design variable | Initial value /mm | Range/mm |
|---|---|---|---|---|---|
| $y_1$ | 52 | (30, 62) | $y_8$ | 48 | (22, 58) |
| $y_2$ | 50 | (26, 60) | $y_9$ | 46 | (18, 56) |
| $y_3$ | 48 | (22, 58) | $y_{10}$ | 44 | (14, 54) |
| $y_4$ | 46 | (18, 56) | $x_1$ | 120 | (60, 200) |
| $y_5$ | 44 | (14, 54) | $x_2$ | 270 | (110, 395) |

| $y_6$ | 52 | (30, 62) | $r_1$ | 10 | (10, 40) |
| $y_7$ | 50 | (26, 60) | $r_2$ | 10 | (5, 40) |

### 4.2 Structures subject to dynamic loading

Under the dynamic loading conditions, two vehicular components are studied, i.e. the thin-walled S-shaped beam (ShB) discussed in our previous studies [96] and thin-walled octagonal multi-cell tube (OMcT) reported in Bai et al [97]. Their geometry parameterization and FE models are illustrated in Figure 3. As critical energy-absorbing parts on the passenger car, these structures can sustain large plastic deformation and dissipate a large amount of kinetic energy due to impact.

In Figure 3(a), the shape of ShB is fully described by 7 design variables and its total length is 1,000 mm. Figure 3(a) also shows the FE model of ShB, which is subjected to the frontal impact at 10 m/s. The specific energy absorption (SEA) is set as the design objective in Equation (11),

$$SEA = \frac{\int_0^d F(x)dx}{M}, \tag{11}$$

where $M$ is the structural mass; $F(x)$ and $d$ are the impact force-displacement history and total deflection, respectively.

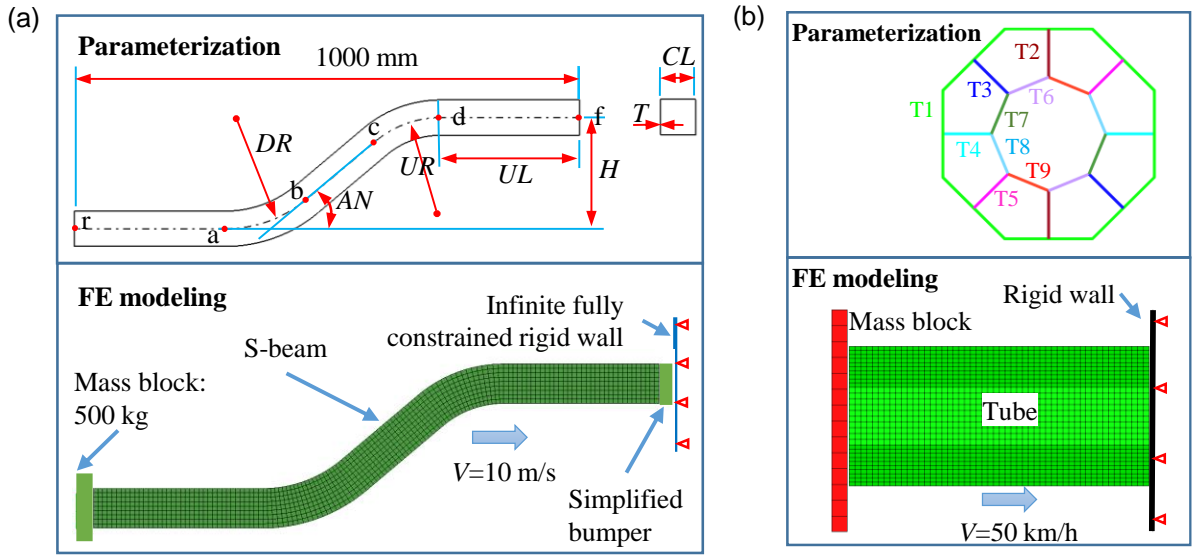

Figure 3 Two structures under dynamic loading: (a) the vehicle frontal S-shaped side beam (ShB) and (b) the octagonal multi-cell tube (OMcT), together with design variables and FE models

In the OMcT model, the cross-section is composed of the inner and outer octagons with ribs. The edge sizes of inner and outer octagons are 30 and 60 mm, respectively. The total length of OMcT is 310 mm. Different colors in Figure 3(b) represent different thicknesses. Each thickness is considered as one design variable and therefore, there are 9 design variables in total. The OMcT FE model impacts a fully constrained rigid wall (velocity: 30 km/h) with the rear end attached to a mass block (600 kg) to represent the vehicle inertia effect. The coefficient force efficiency (CFE) is calculated as the response to assess the energy-absorbing efficiency in Equation (12),

$$CFE = \frac{\int_0^d F(x)dx}{d \times F_{max}},$$ (12)

where $\int_0^d F(x)dx \big/ d$ is the averaged crushing force and $F_{max}$ is the peak crushing force. The detailed material parameters, loading, and boundary conditions can be seen in [96]. These two models were validated in these studies and the details are not repeated here. In the current study, the impact speed of OMcT is increased to 50 km/h to represent the loading condition defined in NCAP standard (National Highway Traffic Safety Administration: https://www.nhtsa.gov/laws-regulations). The four models were simulated by the FE simulation using implicit (TbPT and TqA) and explicit (ShB and OMcT) solvers, that is, the ANSYS Workbench (Ansys, Inc., Workbench, Canonsburg, PA) and LS-DYNA (Ansys, Inc., LSTC, Livermore, CA), respectively. For each structure, 1000 design cases are generated for MLA learning. This same dataset will be used as the training dataset for all of the four MLAs.

## 5. Results

### 5.1 Pareto front of hyperparameters

The Pareto fronts are generated through the SMBO as shown the supplement file. Although the optimal hyperparameters may vary for different datasets, the values of hyperparameters in Pareto front would be close for the datasets with similar sample sizes and design space dimensions [23, 98]. In other words, the optimal hyperparameter values obtained from this study can be applied to the other similar design problems without significant change or used as the basis for optimization [21].

From the Pareto front of multi-objective optimization, a solution, representing a set of hyperparameter values, should be determined as the final optimum considering the articulation of the user's preference [99]. In this study, the posteriori method [100] is used to evaluate each point on the Pareto front after it is generated. By excluding the alternatives with relatively high RMSE or MAXE loss, we select a solution randomly from the remained solutions due to their similar performance. This random selection is used to ensure data generality. A comparison is then made between the models trained by the initial values and selected solutions. For each Pareto front, one hyperparameters group is selected as summarized in Table 4, where corresponding mean loss values of five cross-validations are also included. It needs to be noted that in practical selection for a highly accurate ML model, the training cost and model complexity should be considered.

Table 4 Selected optimal hyperparameters for ML models with respect to the four structural datasets

| MLA | Models | Hyperparameters | | | | | | | Loss | |
|-----|--------|------|------|------|------|------|------|------|------|------|
| | | Kernels | σ | Degree | scale | offset | | | RMSE | MXAE |
| GPR | TbPT | polydot | NA | 3 | 7.22 | -2.24 | | | 0.0603 | 0.3786 |
| | TqA | polydot | NA | 2 | 1.73 | 1.07 | | | 0.0517 | 0.4594 |
| | ShB | polydot | NA | 3 | 2.30 | 1.08 | | | 0.0413 | 0.2341 |
| | OMcT | polydot | NA | 3 | 7.67 | 9.30 | | | 0.0551 | 0.2191 |
| | | C | ε | Kernels | σ | Degree | scale | offset | RMSE | MXAE |
| SVM | TbPT | 0.81 | 0.07 | polydot | NA | 7 | 2.73 | 5.87 | 0.0626 | 0.3777 |
| | TqA | 2.81 | 0.41 | polydot | NA | 1 | 4.03 | -2.09 | 0.0545 | 0.4953 |
| | ShB | 9.77 | 0.13 | laplacedot | 0.37 | NA | NA | NA | 0.0466 | 0.3238 |
| | OMcT | 9.20 | 0.05 | polydot | NA | 2 | 9.25 | 2.96 | 0.0730 | 0.2576 |
| | | Trees | NF | Min TS | | Max TN | | | RMSE | MXAE |
| RFR | TbPT | 773 | 10 | 1 | | 304 | | | 0.0463 | 0.4169 |
| | TqA | 569 | 9 | 1 | | 682 | | | 0.0953 | 0.4414 |
| | ShB | 718 | 7 | 1 | | 1,000 | | | 0.0454 | 0.2975 |
| | OMcT | 879 | 9 | 1 | | 493 | | | 0.0795 | 0.3768 |
| | | Hidden neurons | Activ-ation | Optimizer | Batch size | Learning rate | Momentum | | RMSE | MXAE |
| ANN | TbPT | 26 | relu | sgd | 199 | 0.77 | 0.83 | | 0.0473 | 0.2945 |
| | TqA | 19 | relu | adagrad | 108 | 0.57 | NA | | 0.0506 | 0.4503 |
| | ShB | 8 | tanhdot | adagrad | 85 | 0.30 | NA | | 0.0405 | 0.2571 |
| | OMcT | 36 | tanhdot | sgd | 98 | 0.97 | 0.92 | | 0.0421 | 0.1585 |

*5.2 Effect of hyperparameters optimization*

By using the values in Table 4, sixteen models are trained with 5-folds cross-validation. The distributions of RMSE and MXAE among 5-folds cross-validation are presented by boxplots in Figure 4 and Figure 5, respectively. The results before (trained using initial values in Table 2) and after the HOpt are compared. The median is represented by the black line in the boxes. A lower median value indicates a higher median prediction accuracy. The upper and lower bounds of the box represent the interquartile range with the 25th percentile lower and 75th percentile upper limits. The distance between the upper and lower bounds measures the robustness of the model. Smaller distance indicates higher robustness. For points outside the range of the upper and lower bounds of whiskers are outliers. To further validate the initial and final models, a new dataset is generated by LHS, which includes 180 new designs that are not in the original cross-validation dataset. The distributions of the new test datasets are compared with the corresponding CV datasets in Appendix B to show their consistency. The test errors using the newly generated, unseen data are calculated by the model trained by all original training data in one shot and plotted as the diamond points in Figure 4 and Figure 5. The error values in all three scenarios are listed in Table 5.

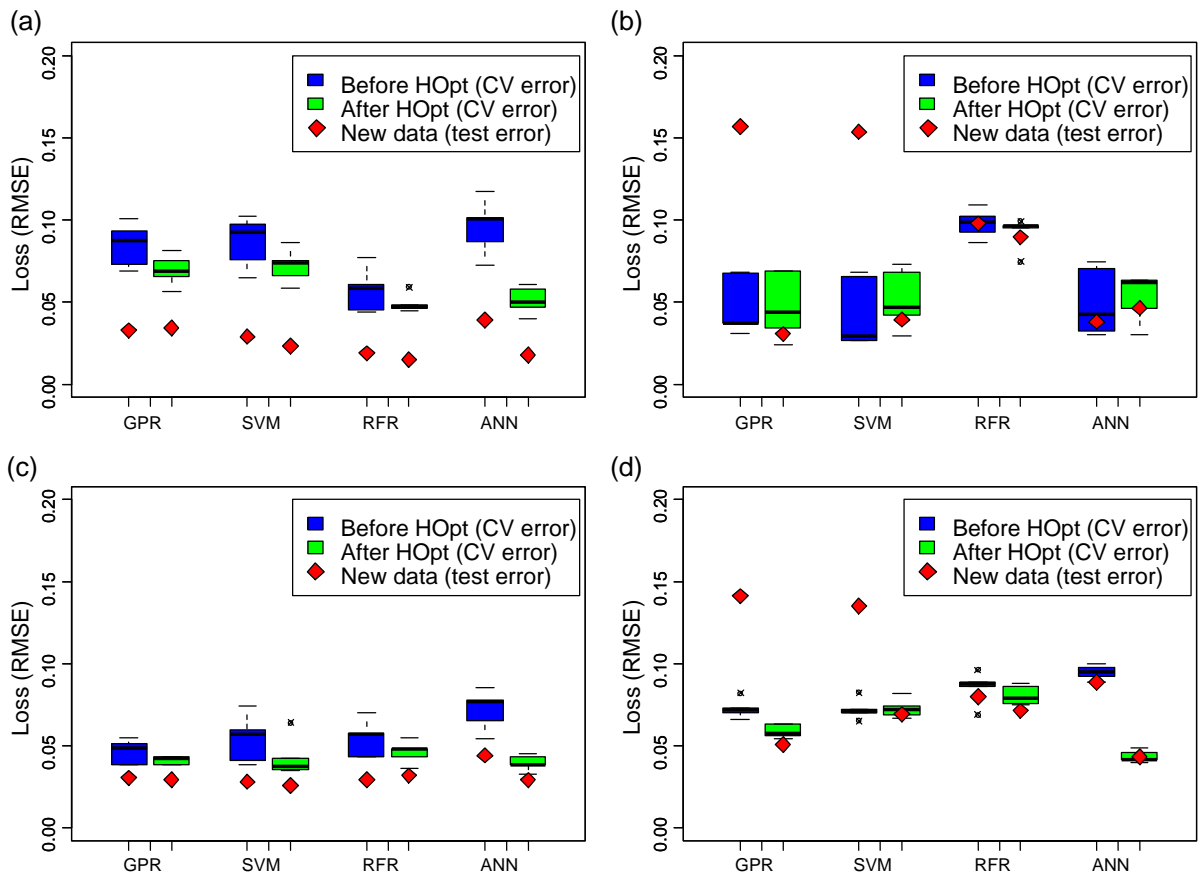

Figure 4 Comparison of the RMSE accuracy of the four MLAs trained by the four structural datasets before and after HOpt , that is, (a) TbPT, (b) TqA, (c) ShB, and (d) OMcT (Note: CV: cross validation)

As shown in Figure 4 and Figure 5, HOpt reduces the median of the RMSE and MXAE values, which indicates the improvement of the prediction accuracy in most case studies with original cross validation (CV) datasets. This is also verified by the new data validation errors. The results of the new dataset show different degrees of performance improvement for various design problems with the same initial hyperparametric values. HOpt brings more evident improvement on the accuracy of TqA and OMcT when GPR and SVM algroithms are used. The errors of TbPT and ShB models can also be reduced, but less siginifcantly, compared to the other cases. In general, HOpt improves the model performance but the degree of improvement depends on specific problems with various complexity.

Similar trends can be observed in model robustness. HOpt can improve the robustness of the MLA models in most cases. However, exceptions can be observed in TqA and OMcT, where the robustness of GPR, SVM, and RFR deteriorates after HOpt. In summary, the initial values of the hyperparameters in most of the machine learning tools are usually determined based on prior knowledge or trial-and-error as summarized in Table 1. The comparisons shown in Figure 4 and Figure 5 indicate that through HOpt, the accuracy and robustness of the ML models can be improved in the mostly used surrogate models in structural design.

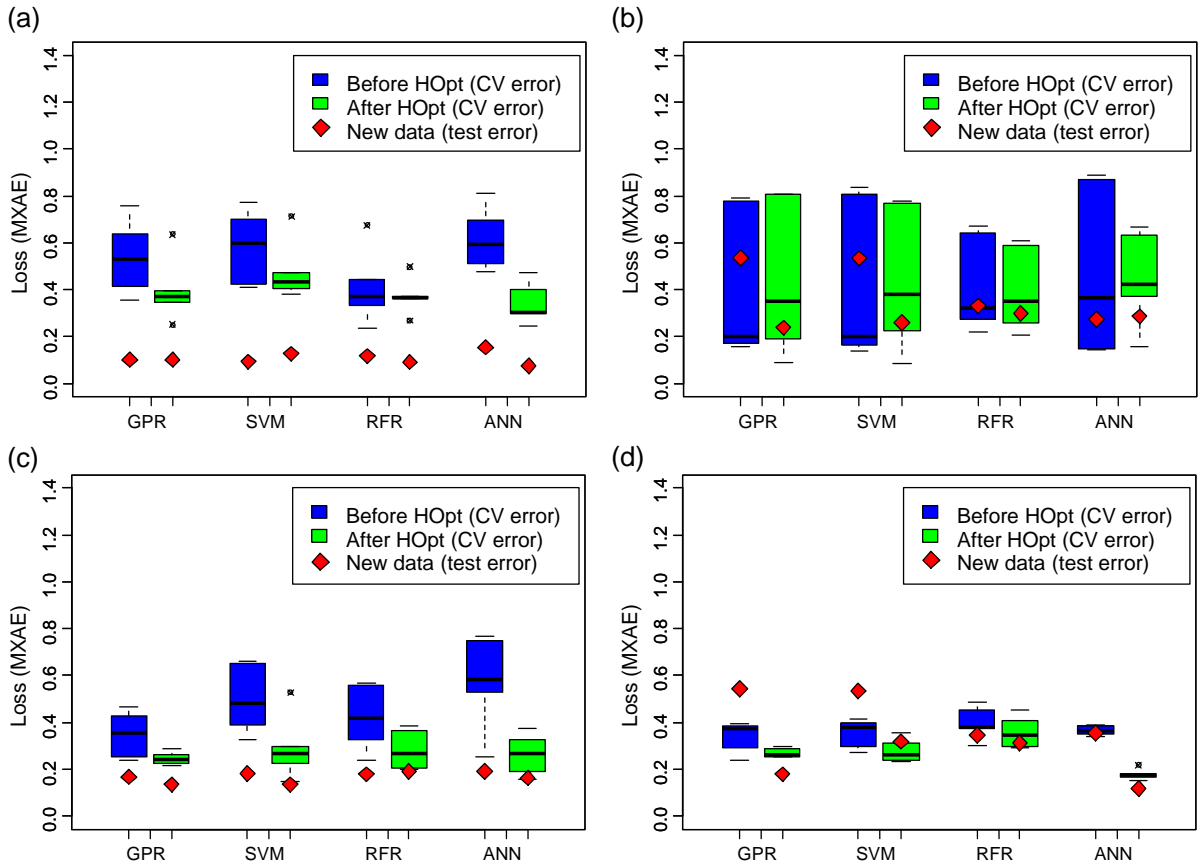

Figure 5 Comparison of the MXAE accuracy of the four MLAs trained by the four structure datasets before and after HOpt, that is, (a) TbPT, (b) TqA, (c) ShB, and (d) OMcT (Note: CV: cross validation)

Besides, in Figure 4 (a) and (c) (also Figure 5 (a) and (c)), the discrepancy between the CV errors of HOpt and new dataset test errors is believed to be caused by different training processes, although the original and new datasets have similar distributions. In the HOpt process, 5-fold CV was used to determine the hyperparameters. In the later training with the new dataset to test the effect of hyperparameters values, no CV is conducted. All the 1,000 data points in the original dataset are applied to conduct a "one-shot" training so the model can use all training data in one time to fit the design problem better. Then, this better-trained model is tested by the 180 new designs and thus the error is reduced in these particular design problems.

In Figure 4 and Figure 5, the results indicate that the performances of some particular ML algorithms (e.g., GPR and SVM) are not stable for specific design problems without a HOpt. When tested using new data, the errors may be increased before HOpt. The degree of instability may vary for different algorithms, hyperparameters, and design problems.

In addition, it is not unusual that the optimization of ML algorithms does not always improve their performance, e.g., the ANN in Figure 4(b), since the results rely on many factors such as dataset size, nonlinearity, and randomity, etc. Here, we do not intend to optimize a specific algorithm for a particular design problem. Instead, we would like to analyze the influence of HOpt on different structural design problems. The results in Table 5 indicate that HOpt can improve the model performance in general.

Table 5 The validation error of MLAs models using the new dataset with error reduction (%) after HOpt

| Cases | Measure | GPR | | | SVM | | | RFR | | | ANN | | |
|-------|---------|-----|---|---|-----|---|---|-----|---|---|-----|---|---|
| | | Before HOpt | After HOpt | Reduction (%) | Before HOpt | After HOpt | Reduction (%) | Before HOpt | After HOpt | Reduction (%) | Before HOpt | After HOpt | Reduction (%) |
| TbPT | RMSE | 0.0327 | 0.0345 | 5.7 | 0.0291 | 0.0234 | -19.8 | 0.0188 | 0.0148 | -21.0 | 0.0391 | 0.0179 | -54.2 |
| | MXAE | 0.1020 | 0.1015 | -0.5 | 0.0925 | 0.1265 | 36.9 | 0.1172 | 0.0919 | -21.6 | 0.1551 | 0.0759 | -51.0 |
| TqA | RMSE | 0.1568 | 0.0306 | -80.5 | 0.1537 | 0.0395 | -74.3 | 0.0979 | 0.0896 | -8.5 | 0.0381 | 0.0463 | 21.7 |
| | MXAE | 0.5370 | 0.2383 | -55.6 | 0.5343 | 0.2600 | -51.3 | 0.3303 | 0.2975 | -9.9 | 0.2748 | 0.2872 | 4.5 |
| ShB | RMSE | 0.0303 | 0.0290 | -4.2 | 0.0280 | 0.0258 | -8.1 | 0.0292 | 0.0317 | 8.6 | 0.0439 | 0.0290 | -34.0 |
| | MXAE | 0.1670 | 0.1339 | -19.8 | 0.1789 | 0.1340 | -25.1 | 0.1783 | 0.1885 | 5.7 | 0.1901 | 0.1619 | -14.8 |
| OMcT | RMSE | 0.1411 | 0.0506 | -64.1 | 0.1351 | 0.0691 | -48.8 | 0.0797 | 0.0712 | -10.7 | 0.0888 | 0.0430 | -51.6 |
| | MXAE | 0.5424 | 0.1777 | -67.2 | 0.5346 | 0.3173 | -40.6 | 0.3448 | 0.3112 | -9.7 | 0.3554 | 0.1166 | -67.2 |

*5.3 Computational time evaluation*

By comparing the time of a single training before and after the HOpt, the impact of HOpt on computational cost is evaluated. The computational power is as follows: Dell Precision Tower 5810 with Intel Xeon CPU E5-2690 v3: 2.6 GHz turbo up to 3.5 GHz and 32 GB RAM. Figure 6 shows the computational time of the four MLAs trained by the four structural datasets before and after HOpt. The results indicate that there is almost no time cost change for GPR training since its hyperparameters have no influence on the model size and then, training task scale. According to Figure 6(d), however, the SVM training time is increased greatly for the OMcT. Regarding this model in Table 4, the large error penalization $C$ (i.e., 9.2) gives a large weight to the error term in Equation (5). Meanwhile, the low value (i.e., 0.05) of $\varepsilon$-bounds causes more samples falling out of $\varepsilon$-bounds. Then, their error with respect to the $\varepsilon$-bounds are added into the error term, which increases the complexity of objective function and the difficulty to minimize errors. These lead to a high training cost.

The training time of RFR is the time used to evaluate the splitting nodes of all trees. In this way, the time of RFR training is closely related to the number of trees (Trees) and NF. Min TS and Max TN can also affect the number of non-leaf nodes since they are related to the single tree scale and then the RFR model size. However, the degree of their influences on the final training time is less compared with the Trees and NF. A larger number of trees and more NF can improve prediction accuracy while they increase the model scale and then the training time.

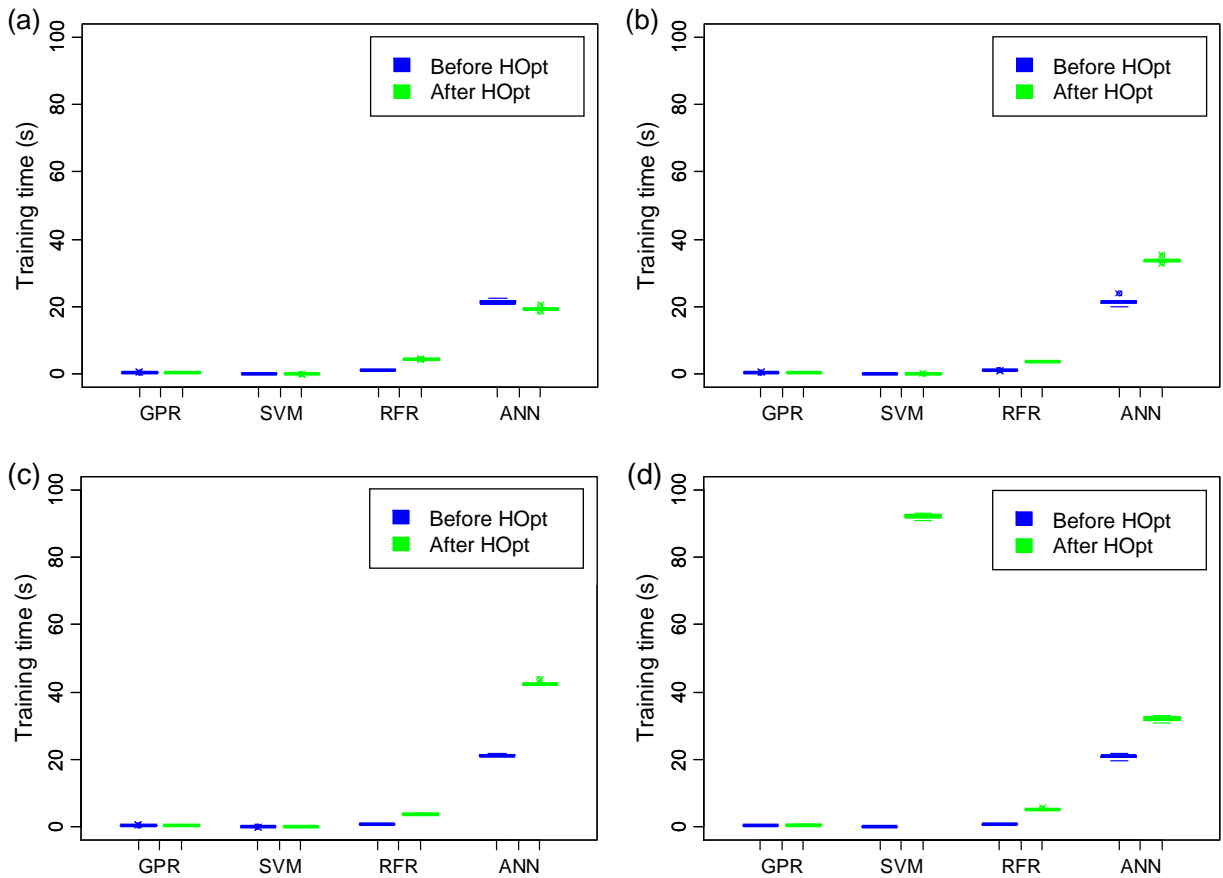

Figure 6 Training time comparison of the 16 models before and after HOpt, that is, (a) TbPT, (b) TqA, (c) ShB, and (d) OMcT

ANN training time is related to the combined effect of the mini-batch size, learning rate, and optimizer and so on. As indicated in Table 4 and Figure 6, after HOpt, the ANN trained by TbPT is reduced slightly, which is caused by the larger learning rate, which accelerates the training convergence. however, the other structural cases presented lower convergence speed although they have larger learning rates, which are caused by the smaller mini-batch sizes. This increases the number of batch training within a single epoch, which rises the training time of a single epoch but lowers the number of epochs for convergence. These suggest that the mini-batch size takes the primary effect on the training time and is followed by the learning rate under a fixed training epoch. The other hyperparameters are more related to model accuracy than the training cost.

In summary, compared with GPR and SVM, the training time of ANN and RFR is more sensitive to hyperparameter values. HOpt improved the ANN performance most significantly, which is followed by the GPR. Therefore, it is recommended to conduct HOpt for ANN if enough computational sources are available. Otherwise, the GPR is recommended for applying the HOpt. Besides, in the HOpt for SVM, small $\varepsilon$-bounds or large error penalization $C$ should be avoided, because they will cause a too complex training objective function and then unexpected singularities and a high convergence cost.

## 6. Discussion

In this section, an analysis is carried out on the relationship between the performance of HOpt method to improve the MLA modeling performance and characteristics of the design problems, as summarized in Table 6. The characteristics include the structure type, the types and number of design variables, load type, and the design domain continuity, which performances are mainly RMSE, MXAE and computational time. Previous studies have demonstrated that the dimensionality and nonlinearity of design problems, which partially depends on the structure and load types, have a strong impact on the surrogate model accuracy and robustness [16, 101]. In addition to structure and load types, the dimensionality of the design space, continuity of the feasible domain, and the design variable type also influence the complexity of the problem. Due to the design constraints, the feasible domains of TbPT and TqA are broken into discontinuous subdomains. The nonlinear boundaries of discontinuous feasible domains pose an additional level of challenge to MLA model training. Furthermore, mixed-variable problems are more challenging for MLA models.

In Table 6, the improvement in accuracy (mean) and robustness (SD) of each MLA model after HOpt are listed for each benchmark problem. The improvement is presented as the percentage of reduction in the values of mean and SD values, where the minus values indicate the reduction of measures and improvement and so versa. The reduction in training time is provided as well. The detailed values of each measure before and after HOpt are listed in the supplement file. In Table 6, the cases with no improvement are denoted with red and the increased training time is denoted with blue. With all the information on problem complexity and MLA performances, we discuss the effect of HOpt in the following three aspects.

- Effect of HOpt on MLA model accuracy

HOpt may not always improve the model accuracy, which depends on specific problems with various complexity. ANN performance for TqA is even deteriorated slightly after HOpt as indicated by CV errors and verified by the new data validation. These were caused by the mixed-variable design space, and we also notice that this case study has the highest dimensionality in all cases. Therefore, the HOpt can improve the MLA-based surrogate model performance in general, but may exhibit unstable performance for high complex problems (e.g., TqA).

Table 6 The characteristics and HOpt performance of the four design benchmark structures

| Category | Item | | Structure | | | |
|---|---|---|---|---|---|---|
| | | | TbPT | TqA | ShB | OMcT |
| Complexity matrix | Structure type | | Truss | Block | Thin-wall | Thin-wall |
| | No. of variables | | 10 | 14 | 7 | 9 |
| | Variables type[*a] | | Single | Mixed | Mixed | Single |
| | Load type | | Static | Static | Impact | Impact |
| | Design domain[*b] | | Discontinuous | Discontinuous | Continuous | Continuous |
| MLA | Measure | | Reduction by % after HOpt ('-' indicates measure reduction or improvement) | | | |
| GPR | RMSE | Mean | -18.8 | 0.0 | -10.9 | -19.2 |
| | | SD | -30.8 | 11.1 | -75.0 | -33.3 |
| | MXAE | Mean | -25.6 | 7.1 | -29.2 | -23.9 |
| | | SD | -12.3 | 2.7 | -70.3 | -34.8 |
| | Time/s | Mean | -19.5 | -21.2 | -15.8 | 17.0 |
| SVM | RMSE | Mean | -16.3 | 18.6 | -20.4 | 0.0 |
| | | SD | -37.5 | -18.2 | -20.0 | -16.7 |
| | MXAE | Mean | -17.5 | 4.0 | -41.9 | -20.5 |
| | | SD | -16.7 | -12.2 | -5.3 | -20.3 |
| | Time/s | Mean | -37.0 | 233.3 | -8.0 | 143931.3 |
| RFR | RMSE | Mean | -14.0 | -5.2 | -14.8 | -5.8 |
| | | SD | -57.1 | 11.1 | -36.4 | -40.0 |
| | MXAE | Mean | -9.4 | -4.9 | -32.5 | -10.1 |
| | | SD | -50.3 | -13.6 | -39.6 | -2.8 |
| | Time/s | Mean | 299.8 | 222.8 | 398.7 | 491.8 |
| ANN | RMSE | Mean | -45.2 | 2.0 | -45.8 | -44.2 |
| | | SD | -46.7 | -38.1 | -54.5 | 0.0 |
| | MXAE | Mean | -44.1 | -6.2 | -53.5 | -42.7 |
| | | SD | -36.8 | -44.5 | -57.5 | -29.4 |
| | Time/s | Mean | -9.5 | 55.4 | 98.5 | 53.2 |

Note: [a] Type of variables:" Single" indicates that only one type of design variables is contained, for example, the nine thickness variables of OMcT. "Mixed" indicates that multiple types of design variables are contained, for example, the radius, length, and node location variables of TqA. [b] Design domain (continuity): continuous or discontinuous.

- Effect of HOpt on MLA model robustness

HOpt improves the robustness of MLA models in the design problems with an intermediate complexity (TbPT and ShB). For the simplest problem (OMcT), HOpt is also helpful, but the improvement is less significant because MLA models already achieve good robustness without HOpt. It should be noted that HOpt may reduce model robustness in a relatively complex design problem (e.g., TqA). This issue can be partially avoided by simplifying a complex problem through dimension reduction and eliminating mixed variables [59, 102].

- Effect of HOpt on MLA training cost

No clear trend can be observed in the relation between HOpt and the training cost. As discussed in Section 5.3, the training time is independent of the features of the design problem but related to the MLA model architecture. The knowledge of MLA model architecture could help estimate and reduce the training time of MLA and the associated HOpt.

To summarize based on the results of present study, HOpt may be unstable when handling design problems with relatively high complexity, but it can improve the modeling performance with less design variables. However, further reduction of the dimension will eliminate the need of HOpt. We recommend GPR as a good start for complex problems due to its low training cost and relatively high accuracy. It is also noticed that HOpt leads to insignificant improvement in simple design problems.

**Conclusions**

In this study, the effects of hyperparameters on MLA surrogate models, namely GPR, SVM, RFR, and ANN are analyzed in detail. A HOpt framework is proposed for optimizing the hyperparameter values to improve model accuracy and robustness. By applying the surrogate modeling and HOpt methods to four benchmark examples, we investigated the impact of HOpt on the accuracy, robustness, and computational costs of the surrogate models.

In general, the HOpt can improve MLA-based surrogate models with lower error. The improvement made by the HOpt depends on the specific problems with various complexity. The performance of HOpt may not be stable when handling complex problems with mixed variables and relative higher-dimensional design space. In addition, GPR is recommended with HOpt due to its low training cost and relatively high accuracy.

The training costs before and after hyperparameters tuning are also evaluated through a single training process. Followed by the RFR, the training time of ANN exhibits the highest sensitivity to the HOpt. ANN training cost is closely related to the learning rate and mini-batch size. The mini-batch size could speed up the convergence and reduce the iteration cost. The RFR training cost is linearly related to the scale of forest (i.e. the number of trees and each tree's nodes). GPR and SVM are insensitive to the hyperparameters tuning. A high computational cost for SVM may occur under the condition of the narrow $\varepsilon$-bounds and the high weight ($C$) of the error term, which should be avoided.

A parametric study conducted on the hyperparameters' influence on RMSE and MXAE is discussed in the supplement file. The results indicate that the polynomial and Laplacian kernels are recommended for the GPR for a high modeling accuracy. The low degree (2 or 3) of the polynomial kernel and low sigma of Laplacian kernel are the good choices for the SVM-based surrogate modeling. Meanwhile, The ANN trained with the *tanh* or *relu* activation functions using *sgd* or *adagrad* optimizers shows a good performance.

In the future work, the codes can be integrated as software with a graphic user interface, and open source will be an option.

**Code repository**

A working repository of this study is available at: https://github.com/Seager1989/HOpt4SMSD.git.

**Acknowledgements**

The first author would like to thank ERAU for the partial financial support through the FIRST grant and Ph.D. program, and the Cooperative Internship Program provided by University of Connecticut and Embry-Riddle Aeronautical University.

**Conflict of Interest**

The authors declare that they have no conflict of interest.

## Appendix A: Convergence histories of HOpt processes for all 16 models

As Figure A.1 shows, all the models have acheived the convergence before the termination criterion (100 evaluations) is reached.

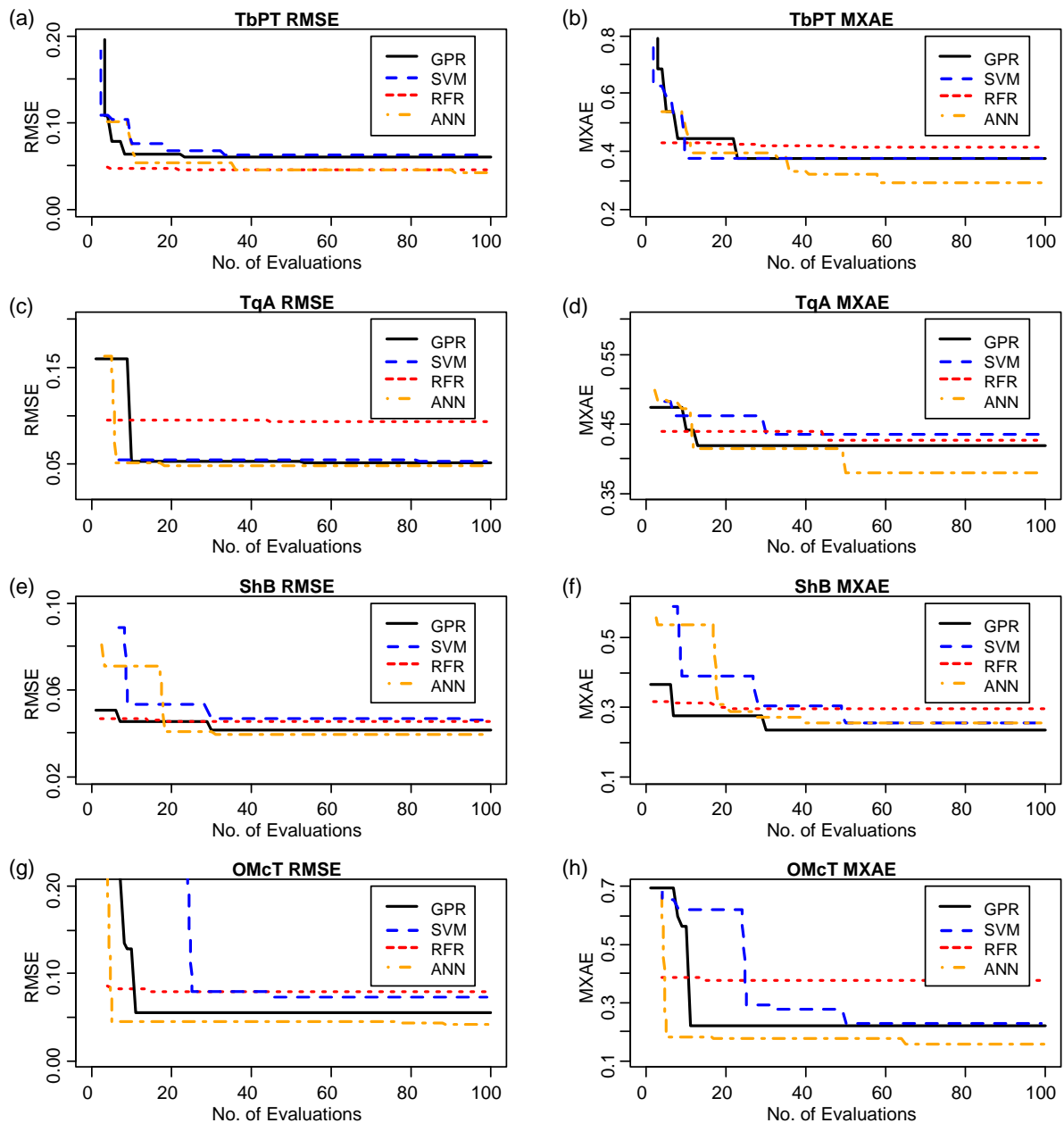

Figure A.1 Convergence histories of the HOpt processes for all the model with 100 evaluations to ensure a convergence

## Appendix B: Response distributions of datasets

As shown in Figure B. 1, the four newly generated test datasets have almost the same distribution as the original training dataset.

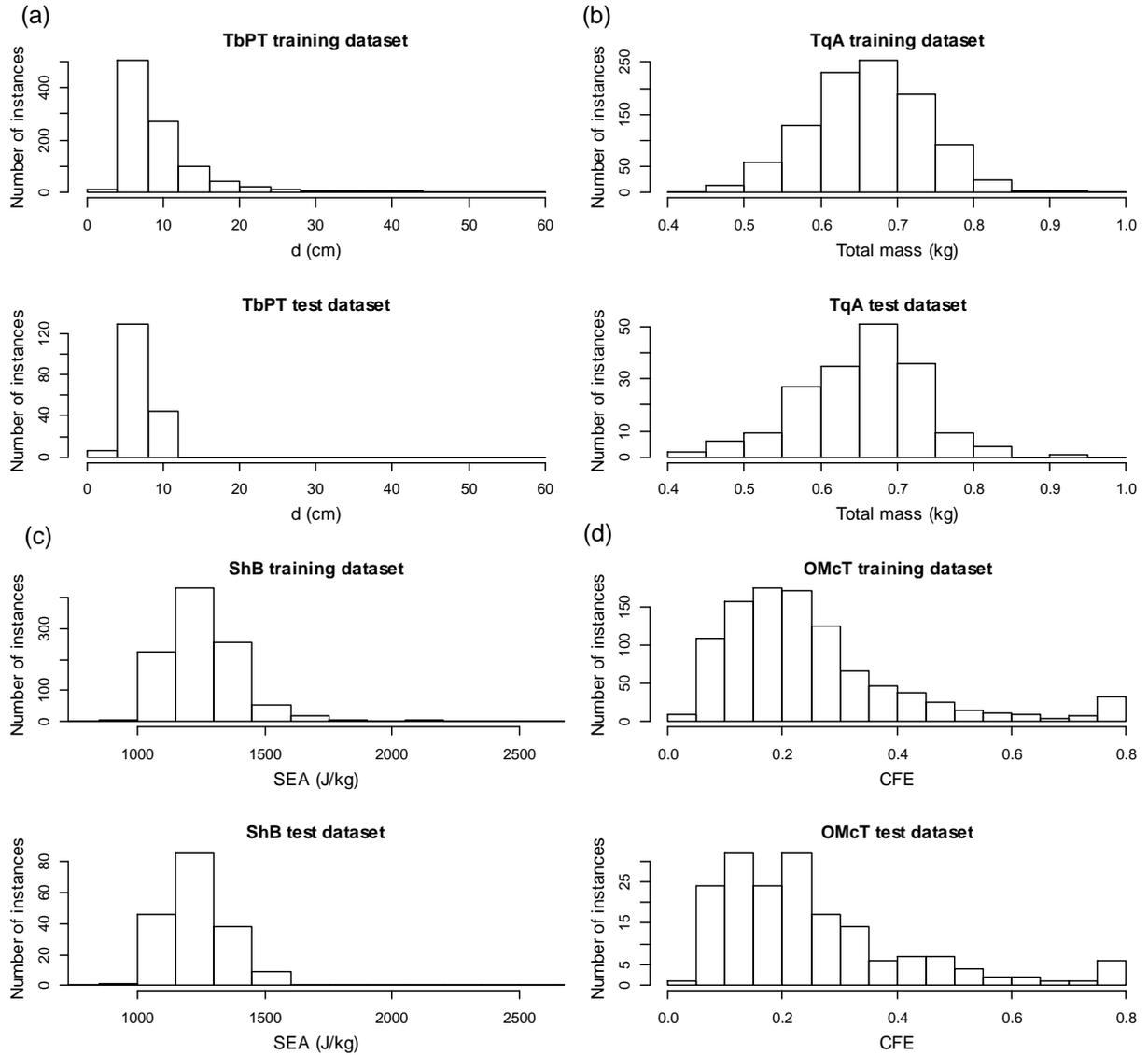

Figure B. 1 Distributions of the training and test datasets of the four structural design cases

# *Supplementary material for:*

# Understanding the effect of hyperparameter optimization on machine learning models for structure design problems

Xianping Du[1+], Hongyi Xu[2], Feng Zhu[1,3*]

*1 Department of Mechanical Engineering, Embry-Riddle Aeronautical University, Daytona Beach, FL 32114, USA*

*2 Department of Mechanical Engineering, University of Connecticut, Storrs, CT 06269, USA*

*3 Hopkins Extreme Materials Institute and Department of Mechanical Engineering, Johns Hopkins University, Baltimore, MD, 21218, USA*

## Supplemental Material Section A:

This supplementary section includes the full table of Pareto fronts from the HOpt.

## Supplemental Material Section B:

This supplementary section includes the hyperparameters effect analysis on the MLA model accuracy (RMSE) for the GPR and ANN based on the four benchmark structures.

## Supplemental Material Section C:

This supplementary section includes the characteristic analysis of the four benchmark problems and corresponding MLA performance before and after the HOpt.

---

[+] Current affiliation: Department of Mechanical and Aerospace Engineering, Rutgers University, Piscataway, NJ 08854, USA

[*] Corresponding author: Hopkins Extreme Materials Institute, Johns Hopkins University, 3400 N Charles St, Baltimore, MD 21218, USA
E-mail: fengzhume@gmail.com (Dr. Feng Zhu); Xianping Du (DUX1@my.erau.edu); Hongyi Xu (hongyi.3.xu@uconn.edu)

Table A.1 All Pareto fronts obtained from the multi-objective HOpt and the frequencies of categorical hyperparameters, together with the average and S.D. of numerical hyperparameters for four structures

| | | Kernel | σ | Degree | Scale | Offset | RMSE | MXAE |
|---|---|---|---|---|---|---|---|---|
| GPR | TbPT | **polydot** | **NA** | **3** | **7.22** | **-2.24** | **0.0603** | **0.3786** |
| | | polydot | NA | 3 | 3.71 | -2.24 | 0.0603 | 0.3786 |
| | TqA | **polydot** | **NA** | **2** | **1.73** | **1.07** | **0.0517** | **0.4594** |
| | | laplacedot | 0.46 | NA | NA | NA | 0.0664 | 0.4189 |
| | | laplacedot | 0.34 | NA | NA | NA | 0.0597 | 0.4208 |
| | | laplacedot | 0.44 | NA | NA | NA | 0.0650 | 0.4193 |
| | | polydot | NA | 2 | 7.93 | 0.37 | 0.0517 | 0.4595 |
| | | polydot | NA | 1 | 1.72 | 5.20 | 0.0537 | 0.4411 |
| | | polydot | NA | 1 | 1.08 | -5.24 | 0.0785 | 0.2932 |
| | ShB | **polydot** | **NA** | **3** | **2.30** | **1.08** | **0.0413** | **0.2341** |
| | | polydot | NA | 3 | 1.20 | 4.15 | 0.0413 | 0.2342 |
| | OMcT | **polydot** | **NA** | **3** | **7.67** | **9.30** | **0.0551** | **0.2191** |
| Statistics | Avg. | polydot: 9 | 0.41 | 2.33 | 3.84 | 1.27 | 0.0571 | 0.3630 |
| | S.D. | laplacedot:3 | 0.05 | 0.82 | 2.76 | 4.17 | 0.0100 | 0.0883 |

| | | C | ε | Kernel | σ | Degree | Scale | Offset | RMSE | MXAE |
|---|---|---|---|---|---|---|---|---|---|---|
| SVM | TbPT | **0.81** | **0.07** | **polydot** | **NA** | **7** | **2.73** | **5.87** | **0.0626** | **0.3777** |
| | TqA | **2.81** | **0.41** | **polydot** | **NA** | **1** | **4.03** | **-2.09** | **0.0545** | **0.4953** |
| | | 2.81 | 0.51 | polydot | NA | 1 | 4.77 | -3.81 | 0.0580 | 0.4925 |
| | | 3.53 | 0.28 | polydot | NA | 1 | 5.60 | 0.08 | 0.0541 | 0.5057 |
| | | 3.71 | 0.12 | polydot | NA | 1 | 5.67 | 0.69 | 0.0534 | 0.5062 |
| | | 3.14 | 0.72 | rbfdot | 0.05 | NA | NA | NA | 0.0754 | 0.4358 |
| | ShB | **9.77** | **0.13** | **laplacedot** | **0.37** | **NA** | **NA** | **NA** | **0.0466** | **0.3238** |
| | | 7.56 | 0.85 | polydot | NA | 4 | 1.11 | 6.61 | 0.0600 | 0.2558 |
| | | 3.26 | 0.86 | polydot | NA | 4 | 2.57 | 5.81 | 0.0600 | 0.2748 |
| | | 9.32 | 0.43 | polydot | NA | 5 | 1.34 | 2.64 | 0.0586 | 0.2856 |
| | | 9.81 | 0.05 | laplacedot | 0.46 | NA | NA | NA | 0.0465 | 0.3295 |
| | OMcT | **9.20** | **0.05** | **polydot** | **NA** | **2** | **9.25** | **2.96** | **0.0730** | **0.2576** |
| | | 7.34 | 0.69 | polydot | NA | 2 | 9.27 | 2.64 | 0.0808 | 0.2231 |
| Statistics | Avg. | 5.62 | 0.40 | polydot: 10 laplacedot:2 | 0.29 | 2.8 | 4.63 | 2.14 | 0.0603 | 0.3664 |
| | S.D. | 3.12 | 0.30 | rbfdot: 1 | 0.18 | 1.99 | 2.76 | 3.29 | 0.0101 | 0.1034 |

| | | Trees | NF | Min TS | Max TN | | RMSE | MXAE |
|---|---|---|---|---|---|---|---|---|
| RFR | | Trees | NF | Min TS | Max TN | | RMSE | MXAE |

| | | | | | | | |
|---|---|---|---|---|---|---|---|
| | TbPT | **773** | **10** | **1** | **304** | **0.0463** | **0.4169** |
| | TqA | **569** | **9** | **1** | **682** | **0.0953** | **0.4414** |
| | | 844 | 1 | 1 | 982 | 0.1152 | 0.4327 |
| | | 496 | 14 | 26 | 346 | 0.1122 | 0.4383 |
| | | 161 | 14 | 1 | 1,000 | 0.0955 | 0.4402 |
| | | 532 | 1 | 7 | 974 | 0.1176 | 0.4274 |
| | | 80 | 14 | 3 | 920 | 0.0957 | 0.4383 |
| | | 457 | 14 | 1 | 1,000 | 0.0947 | 0.4438 |
| | ShB | **718** | **7** | **1** | **1,000** | **0.0454** | **0.2975** |
| | | 415 | 7 | 1 | 475 | 0.0457 | 0.2963 |
| | OMcT | **879** | **9** | **1** | **493** | **0.0795** | **0.3768** |
| | | 153 | 9 | 1 | 590 | 0.0795 | 0.3752 |
| | | 463 | 9 | 1 | 564 | 0.0795 | 0.3756 |
| Statistics | Avg. | 503.1 | 9.1 | 3.5 | 717.7 | 0.0848 | 0.4000 |
| | S.D. | 249.9 | 4.3 | 6.7 | 259.5 | 0.0246 | 0.0506 |

| | | Hidden neurons | Activation | Optimizer | Batch size | Learning rate | Momentum | RMSE | MXAE |
|---|---|---|---|---|---|---|---|---|---|
| ANN | TbPT | **26** | **relu** | **sgd** | **199** | **0.77** | **0.83** | **0.0473** | **0.2945** |
| | | 45 | relu | sgd | 199 | 0.77 | 0.73 | 0.0423 | 0.3186 |
| | TqA | **19** | **relu** | **adagrad** | **108** | **0.57** | **NA** | **0.0506** | **0.4503** |
| | | 97 | relu | adagrad | 146 | 0.43 | NA | 0.0514 | 0.4183 |
| | | 65 | relu | adagrad | 166 | 0.55 | NA | 0.0803 | 0.3799 |
| | | 26 | relu | adagrad | 196 | 0.13 | NA | 0.0490 | 0.4699 |
| | ShB | **8** | **tanhdot** | **adagrad** | **85** | **0.30** | **NA** | **0.0405** | **0.2571** |
| | | 7 | tanhdot | adagrad | 86 | 0.30 | NA | 0.0394 | 0.2682 |
| | | 14 | tanhdot | adagrad | 184 | 0.38 | NA | 0.0395 | 0.2626 |
| | OMcT | **36** | **tanhdot** | **sgd** | **98** | **0.97** | **0.92** | **0.0421** | **0.1585** |
| Statistics | Avg. | 34.3 | relu: 6 | sgd: 3 | 146.7 | 0.52 | 0.83 | 0.0482 | 0.3278 |
| | S.D. | 26.9 | tanhdot: 4 | adagrad: 7 | 45.9 | 0.25 | 0.08 | 0.0115 | 0.0942 |

**Supplemental Material Section B:**

*B.1 Hyperparameters effect on GPR*

The only hyperparameters in GPR are kernel-related parameters, that is, the four kernel functions in Equation 16. The GPR performance with different *Sigma* ($\sigma$) is plotted in Figure B.1 (a) and (b) for radial basis and Laplacian kernel, respectively. The low $\sigma\,(<2)$ level takes a critical effect. The best GPR performance $\sigma$ are 1 and 0 for radial basis and laplacian kernels, respectively.

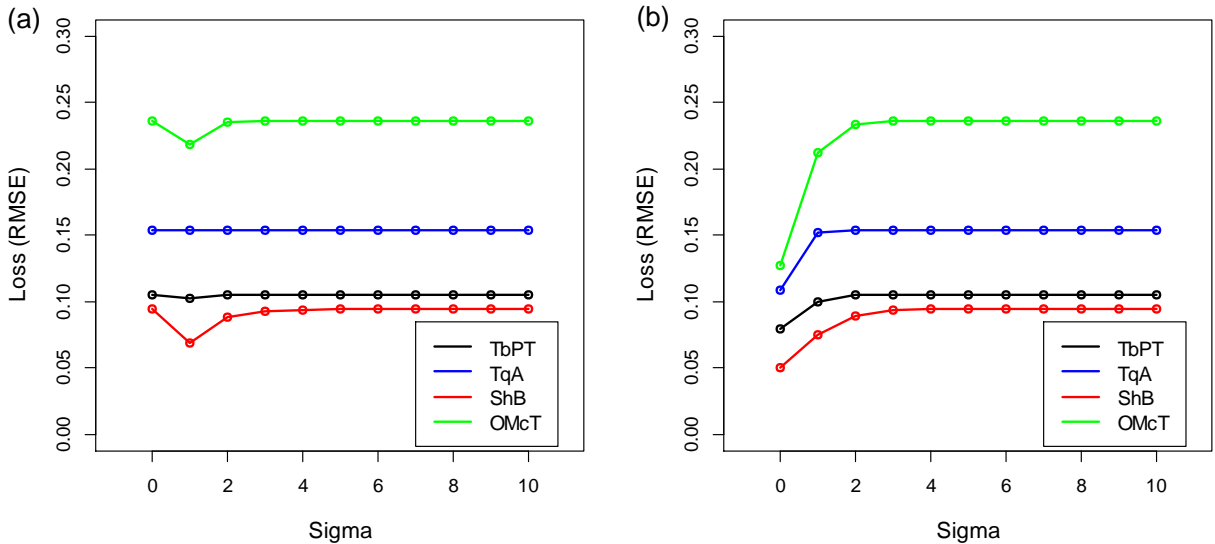

Figure B.1 Effects of kernel hyperparameter *sigma* on the GPR performance for (a) radial basis dot and (b) Laplacian kernels

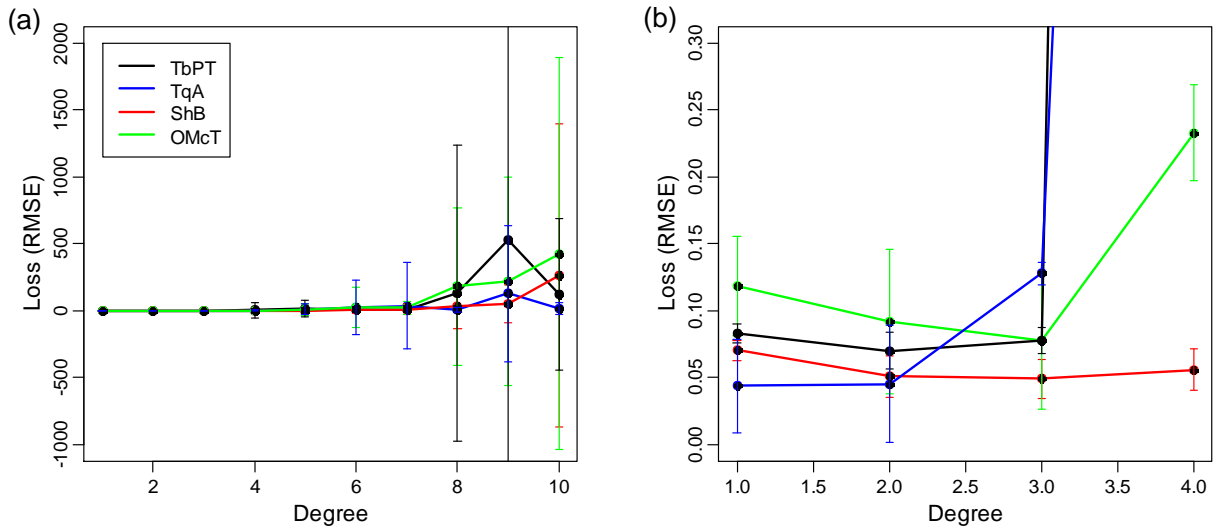

Figure B.2 The effect of polynomial *degree* on GPR performance with the error to account the influence of the *scale* and *offset*: (a) the whole range of *degree* and (b) the enlarged view on the low degree level

Regarding the polynomial kernel, different *degree*s influence model accuracy significantly. Table 4 shows that the optimal degrees would be 2 or 3, so low degree levels in Figure B.2(a) is amplified to see the details in Figure

B.2(b). The 2nd or 3rd order polynomial also presented a high model accuracy. The influence of *scale* and *offset* was increased with the degree, which is indicated by the larger standard errors.

Furthermore, for *scale* and *offset*, their absolute values should be larger than 1 since this will significantly reduce the RMSE error as shown in Figure B.3. If their absolute values are larger than 1, it generally works well to generate an accurate GPR. The hyperbolic tangent kernel also contains the parameters of *scale* and *offset*. Figure B.4 shows their influence on GPR performance, which indicates a similar trend in four cases. The *scale* and *offset* are suggested to be around 1 and 10, respectively, for an accurate GPR.

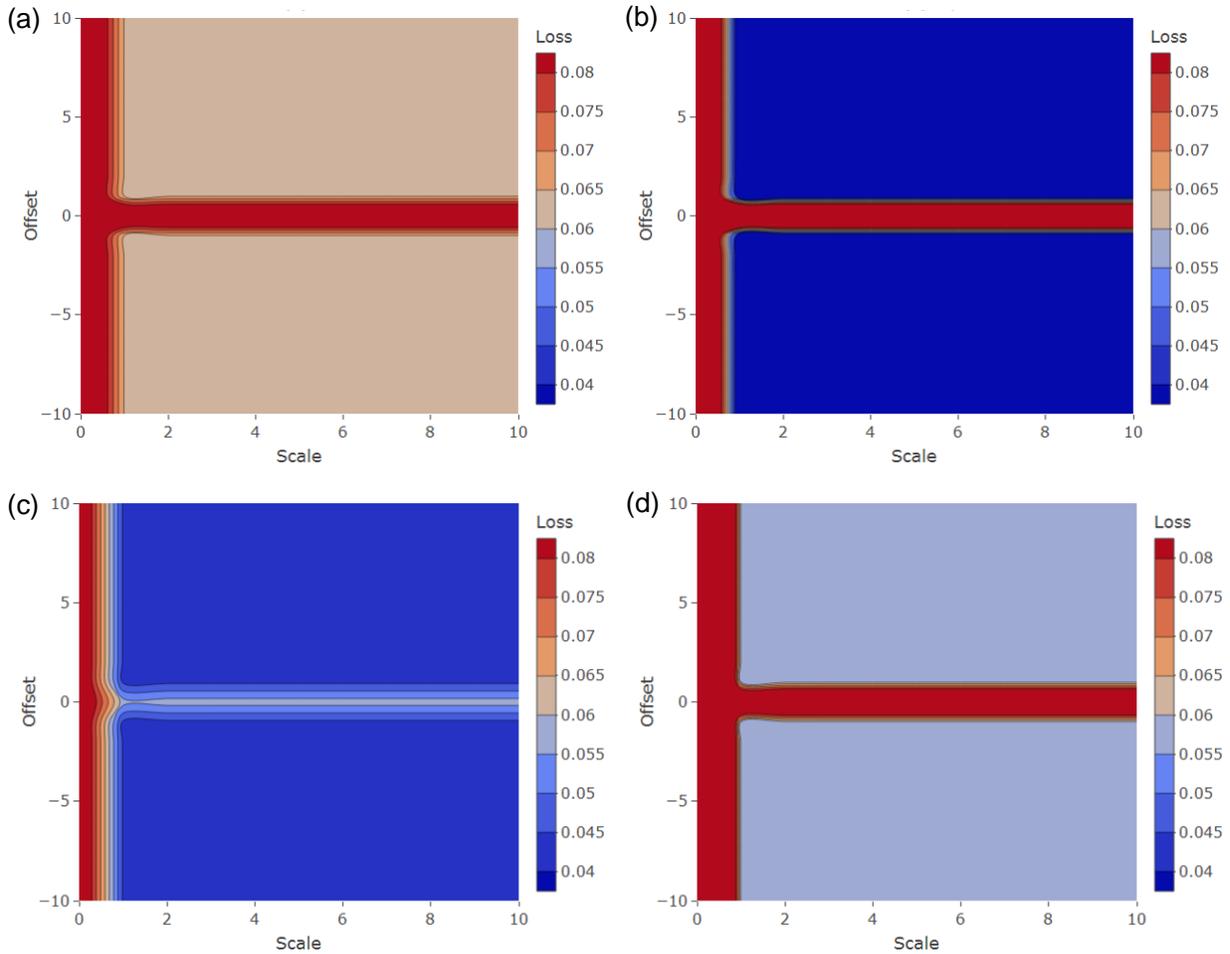

Figure B.3 GPR performance with different *scale* and *offset* under the *degree*s of (a) ShB with degree 3; (b) OMcT with degree 3; (c) TbPT with degree 2; (d) TqA with degree 2

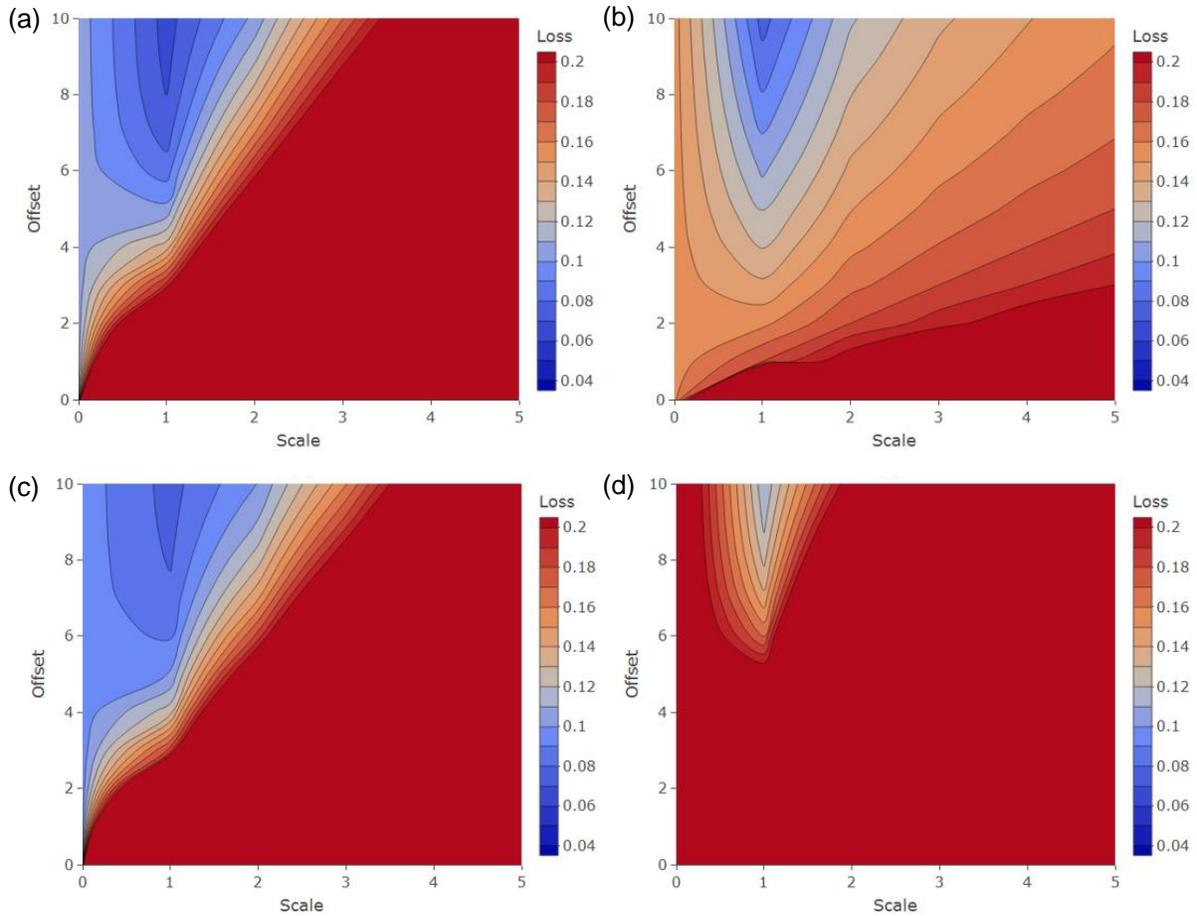

Figure B.4 GPR performance with the change of the hyperbolic tangent kernel function's *scale* and *offset* for (a) ShB; (b) OMcT; (c) TbPT; (d) TqA

In summary, the RMSE accuracy of sixteen models trained with the best kernel hyperparameters values is presented in Figure B.5. The polynomial kernel shows the best performance on GPR accuracy. Meanwhile, the radial basis kernel demonstrates a large scatter and less unstable performance.

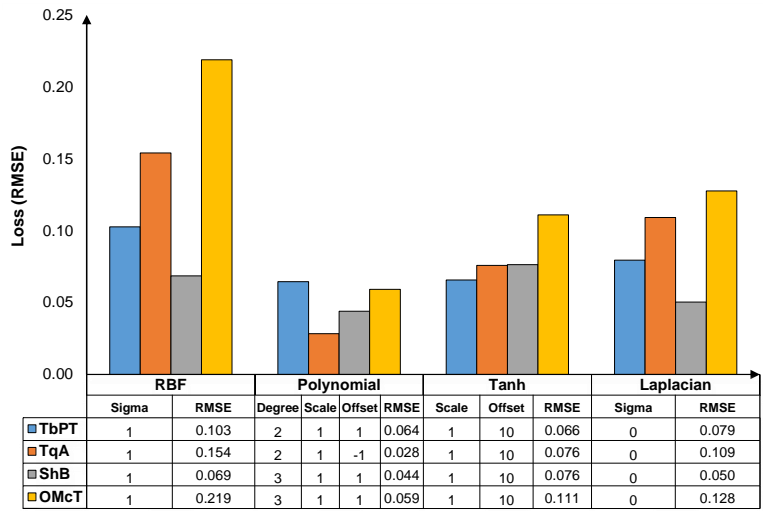

| | RBF | | Polynomial | | | | Tanh | | | Laplacian | |
|---|---|---|---|---|---|---|---|---|---|---|---|---|
| | Sigma | RMSE | Degree | Scale | Offset | RMSE | Scale | Offset | RMSE | Sigma | RMSE |
| TbPT | 1 | 0.103 | 2 | 1 | 1 | 0.064 | 1 | 10 | 0.066 | 0 | 0.079 |
| TqA | 1 | 0.154 | 2 | 1 | -1 | 0.028 | 1 | 10 | 0.076 | 0 | 0.109 |
| ShB | 1 | 0.069 | 3 | 1 | 1 | 0.044 | 1 | 10 | 0.076 | 0 | 0.050 |
| OMcT | 1 | 0.219 | 3 | 1 | 1 | 0.059 | 1 | 10 | 0.111 | 0 | 0.128 |

Figure B.5 Comparision on the modeling performances of the four kernel functions in the GPR algorithm

*B.2 Hyperparameters effect on ANN*

To explore the hyperparameters effect of ANN, a basic one-hidden-layer ANN is constructed with the following hyperparameter values based on the optimization: batch size=100, learning rate=0.1, hidden neurons=20, optimizer=*sgd*, activation function = *tanh*, momentum=0.9. Also, only the ShB dataset is used considering the high computational cost of ANN training and the similar trend of hyperparameters effect for each dataset. All models in this part are trained with 10,000 epochs.

The influence of batch size and the learning rate is studied and plotted with log scale contour in Figure B.6. Larger learning rates and smaller batch sizes speed up the convergence, but too large learning rates could lead to unstable training and may lose the optimum. However, too small mini-batch size could increase the communication cost, i.e. the computational cost to synchronize the shared variables (gradient or model parameters, etc.) between different mini-batches, although it could increase the convergence speed [1]. Base on Figure B.6, a learning rate of 0.1 is determined as the optimum. Under this condition, the larger mini-batch size is preferred only if the same accuracy can be achieved to reduce communication costs [1]. Therefore, the mini-batch size is set as 100 for the current dataset size.

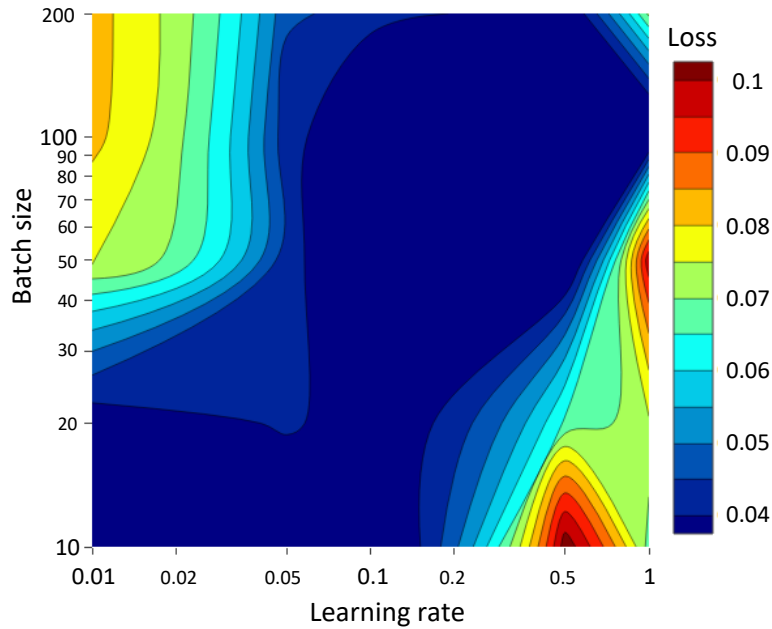

Figure B.6 Mutual influence of the learning rate and mini-batch size on ANN loss (RMSE) in log-scale

The influences of other hyperparameters, i.e. the hidden neuron, optimizer, activation function, and layer structures, on the convergence process are illustrated in Figure B.7. Figure B.7(a) shows that more hidden neurons, indicating more weight factors to be optimized, would slow down the training convergence. Considering the training dataset size (i.e. 1,000) in this part, 20 hidden neurons are used since it results in slightly better accuracy than the 5. In Figure B.7(b), among four optimizers, adagrad results in the highest accuracy, stability, and convergence speed. Although the *sgd* shows a similar accuracy, it is limited by the low convergence speed caused by its complicated training process [2]. Hence, *adagrad* would be a good choice.

In Figure B.7(c), *relu* shows low RMSE accuracy due to its linear behavior in Figure 2. After the transmission through one hidden layer with the linear activation function, the output of ANN will be still linear, which is not suitable for the problems with a non-linear response, such as ShB. Except for *relu*, the other three activation functions are all nonlinear and generated similar accuracy levels. The *tanh* reaches a quick and stable convergence due to its large value variation in the range of [0 1]. Therefore, *tanh* is used for the nonlinear response prediction.

As discussed, a multi-layered ANN possesses a higher ability to learn high nonlinear responses. To explore the multi-layers effect on ANN training, another four multi-layered models are built with optimal hyperparameters identified earlier. As shown in Figure B.7(d), 7-20-1 presents the best performance. The model 7-10-5-1 suggests a good but slightly lower accuracy since its number of weights (i.e. 125) is also less than a quarter of the training data set size. The other three ANN structures greatly increase the number of weights and lead to unstable convergence process and deteriorated accuracy, which indicates insufficient training. This means compared with increasing the number of layers, the number of weights is more important to fully train an accurate ANN.

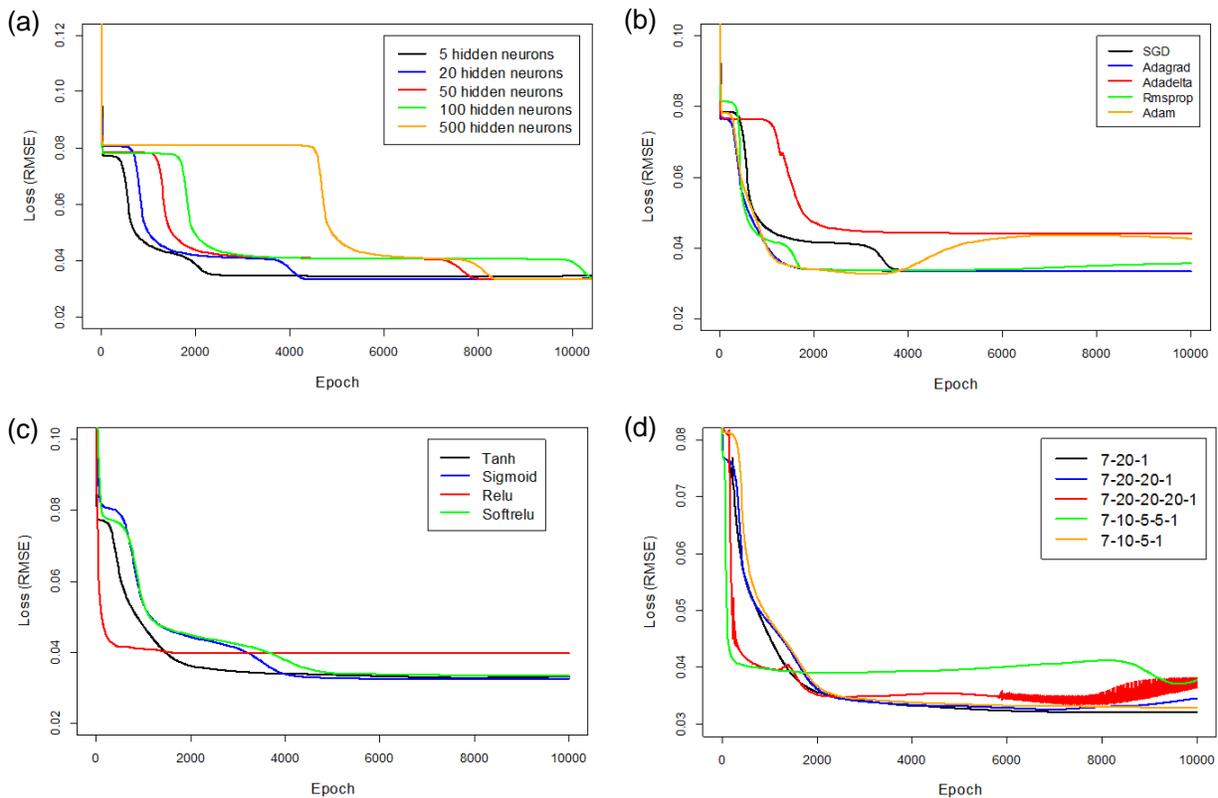

Figure B.7 Loss convergence histories in ANN models with respect to the (a) hidden neurons; (b) optimizers; (c) activation functions and (d) hidden layer structures, where 7-20-1 denotes 7 input features, 20 hidden layer neurons, and 1 output. The other ANN models are denoted in a similar manner.

**Supplemental Material Section C:**

Table C.1 The mean and standard deviation of the values of the measures of the sixteen ML models before and after the HOpt

| Category | Items | | TbPT | | TqA | | ShB | | OMcT | |
|---|---|---|---|---|---|---|---|---|---|---|
| MLA | Measure | | Before\|after HOpt | | Before\|after HOpt | | Before\|after HOpt | | Before\|after HOpt | |
| GPR | RMSE | Mean | 0.085 | 0.069 | 0.048 | 0.048 | 0.046 | 0.041 | 0.073 | 0.059 |
| | | SD | 0.013 | 0.009 | 0.018 | 0.020 | 0.008 | 0.002 | 0.006 | 0.004 |
| | MXAE | Mean | 0.539 | 0.401 | 0.421 | 0.451 | 0.349 | 0.247 | 0.335 | 0.255 |
| | | SD | 0.163 | 0.143 | 0.333 | 0.342 | 0.101 | 0.030 | 0.069 | 0.045 |
| | Time/s | Mean | 0.532 | 0.428 | 0.528 | 0.416 | 0.530 | 0.446 | 0.506 | 0.592 |
| SVM | RMSE | Mean | 0.086 | 0.072 | 0.043 | 0.051 | 0.054 | 0.043 | 0.072 | 0.072 |
| | | SD | 0.016 | 0.010 | 0.022 | 0.018 | 0.015 | 0.012 | 0.006 | 0.005 |
| | MXAE | Mean | 0.583 | 0.481 | 0.430 | 0.447 | 0.501 | 0.291 | 0.352 | 0.280 |
| | | SD | 0.162 | 0.135 | 0.360 | 0.316 | 0.151 | 0.143 | 0.064 | 0.051 |
| | Time/s | Mean | 0.054 | 0.034 | 0.042 | 0.140 | 0.050 | 0.046 | 0.064 | 92.18 |
| RFR | RMSE | Mean | 0.057 | 0.049 | 0.097 | 0.092 | 0.054 | 0.046 | 0.086 | 0.081 |
| | | SD | 0.014 | 0.006 | 0.009 | 0.010 | 0.011 | 0.007 | 0.010 | 0.006 |
| | MXAE | Mean | 0.413 | 0.374 | 0.425 | 0.404 | 0.421 | 0.284 | 0.398 | 0.358 |
| | | SD | 0.165 | 0.082 | 0.214 | 0.185 | 0.144 | 0.087 | 0.072 | 0.070 |
| | Time/s | Mean | 1.102 | 4.406 | 1.158 | 3.738 | 0.786 | 3.920 | 0.904 | 5.350 |
| ANN | RMSE | Mean | 0.093 | 0.051 | 0.050 | 0.051 | 0.072 | 0.039 | 0.077 | 0.043 |
| | | SD | 0.015 | 0.008 | 0.021 | 0.013 | 0.011 | 0.005 | 0.003 | 0.003 |
| | MXAE | Mean | 0.619 | 0.346 | 0.482 | 0.452 | 0.564 | 0.262 | 0.309 | 0.177 |
| | | SD | 0.144 | 0.091 | 0.373 | 0.207 | 0.212 | 0.090 | 0.034 | 0.024 |
| | Time/s | Mean | 21.56 | 19.51 | 21.76 | 33.81 | 21.44 | 42.55 | 20.92 | 32.05 |